%% file: main.tex
\pdfoutput=1

\documentclass[11pt]{article}

\usepackage[final]{acl} 

\usepackage{times}
\usepackage{latexsym}

\usepackage[T1]{fontenc}

\usepackage[utf8]{inputenc}

\usepackage{microtype}
\usepackage{graphicx}
\usepackage{multirow}
\usepackage{caption}
\usepackage{booktabs}
\usepackage{epsfig}
\usepackage{adjustbox}
\usepackage{amsfonts, amsmath, amssymb}
\usepackage{booktabs} 
\usepackage{comment}
\usepackage{caption, subcaption}
\usepackage{textcomp}
\usepackage{relsize}
\usepackage{stmaryrd}
\usepackage{bbm}
\usepackage{rotating}
\usepackage{helvet}
\usepackage{courier}
\usepackage{cleveref}
\usepackage{xspace}
\usepackage{enumitem}
\usepackage{mdframed}
\usepackage{makecell}
\usepackage{tcolorbox}
\usepackage{nicematrix}
\usepackage{tabu}
\usepackage{colortbl}
\usepackage{xcolor}
\usepackage{float} 
\usepackage{array} 
\usepackage{pifont}
\usepackage{tabularx} 
\usepackage{arydshln}
\usepackage{pgfplots}
\pgfplotsset{compat=1.17}
\usepackage{tikz}
\usepackage{xcolor}
\usepackage{booktabs}
\usepackage{multirow}
\usepackage{siunitx}
\newcolumntype{C}[1]{>{\centering\arraybackslash}p{#1}}
\newcolumntype{P}{>{\centering\arraybackslash}m{0.1\linewidth}}

\PassOptionsToPackage{table}{xcolor}

\title{Can Large Language Models be Effective Online Opinion Miners?}

\usepackage{footmisc}
\DefineFNsymbols{mySymbols}{{\ensuremath\dagger}{\ensuremath\ddagger}\S\P
   *{**}{\ensuremath{\dagger\dagger}}{\ensuremath{\ddagger\ddagger}}}
\setfnsymbol{mySymbols}

\author{
    Ryang Heo~~~
    Yongsik Seo~~~
    Junseong Lee~~~
    Dongha Lee\thanks{\; Corresponding author}\\
    Department of Artificial Intelligence, Yonsei University -  DLI Lab \\
    \texttt{\{ryang1119, ysseo, brulee, donalee\}@yonsei.ac.kr}\\   
}

\usepackage{graphicx}
\usepackage{multirow}
\usepackage{pdfpages}

\begin{document}
\maketitle
\input{dfn}
\begin{abstract}
\input{000abstract}
\end{abstract}

\section{Introduction}
\label{sec:intro}

\input{010introduction}

\section{\proposed Benchmark}
\label{sec:data}
\input{020dataset}

\section{Experiments}
\label{sec:task_eval}

\input{030task_eval}


\section{Results and Discussion}
\label{sec:results}
\input{050results}

\section{Related Work}
\label{sec:relwork}
\input{060relatedwork}

\section{Conclusion}
\label{sec:conclusion}

\input{070conclusion}

\section*{Limitations}
\label{sec:limitations}
\input{080limitations}

\section*{Ethical Statement}
\label{sec:etical}
\input{090ethical_statement}

\section*{Acknowledgments}
\label{sec:acknowledge}
\input{100acknowledgements}

\bibliography{reference}

\newpage
\clearpage
\appendix
\label{sec:appendix}
\input{081appendix}

\end{document}

%% file: dfn.tex
\newcommand{\cmark}{\textcolor{green!50!black}{\ding{51}}}
\newcommand{\xmark}{\textcolor{red!75!black}{\ding{55}}} 

\newcommand{\subone}{Feature-centric opinion extraction\xspace}
\newcommand{\subtwo}{Opinion-centric insight generation\xspace}

\newcommand{\proposed}{OOMB\xspace}

\newcommand{\gptthree}{GPT-3.5-turbo\xspace}
\newcommand{\gptmini}{GPT-4o-mini\xspace}
\newcommand{\gptfouro}{GPT-4o\xspace}
\newcommand{\gptfour}{GPT-4-turbo\xspace}
\newcommand{\claudehaiku}{Claude-3.5-Haiku\xspace}
\newcommand{\claudesonnet}{Claude-3.5-Sonnet\xspace}
\newcommand{\llamaeight}{Llama3-8B-Instruct\xspace}
\newcommand{\llamaseventy}{Llama3-70B-Instruct\xspace}
\newcommand{\gemmanine}{Gemma2-9B-it\xspace}
\newcommand{\gemmatwentyseven}{Gemma2-27B-it\xspace}
\newcommand{\qwen}{Qwen2.5-7B-Instruct\xspace}
\newcommand{\deepseek}{DeepSeek-7B-chat\xspace}

%% file: 000abstract.tex
The surge of user-generated online content presents a wealth of insights into customer preferences and market trends.
However, the highly diverse, complex, and context-rich nature of such content poses significant challenges to traditional opinion mining approaches.
To address this, we introduce \underline{\textbf{O}}nline \underline{\textbf{O}}pinion \underline{\textbf{M}}ining \underline{\textbf{B}}enchmark (\textbf{\proposed}), a novel dataset and evaluation protocol designed to assess the ability of large language models (LLMs) to mine opinions effectively from diverse and intricate online environments. 
\proposed provides, for each content instance, an extensive set of \textit{(entity, feature, opinion)} tuples and a corresponding opinion-centric \textit{insight} that highlights key opinion topics, thereby enabling the evaluation of both the extractive and abstractive capabilities of models.
Through our proposed benchmark, we conduct a comprehensive analysis of which aspects remain challenging and where LLMs exhibit adaptability, to explore whether they can effectively serve as opinion miners in realistic online scenarios.
This study lays the foundation for LLM-based opinion mining and discusses directions for future research in this field.
Our code and dataset are available\footnote{\url{https://github.com/ryang1119/OOMB}}.



%% file: 010introduction.tex
\input{TAB/020data_comparison}

\begin{figure}[!t]
    \centering
    \includegraphics[width=\columnwidth]{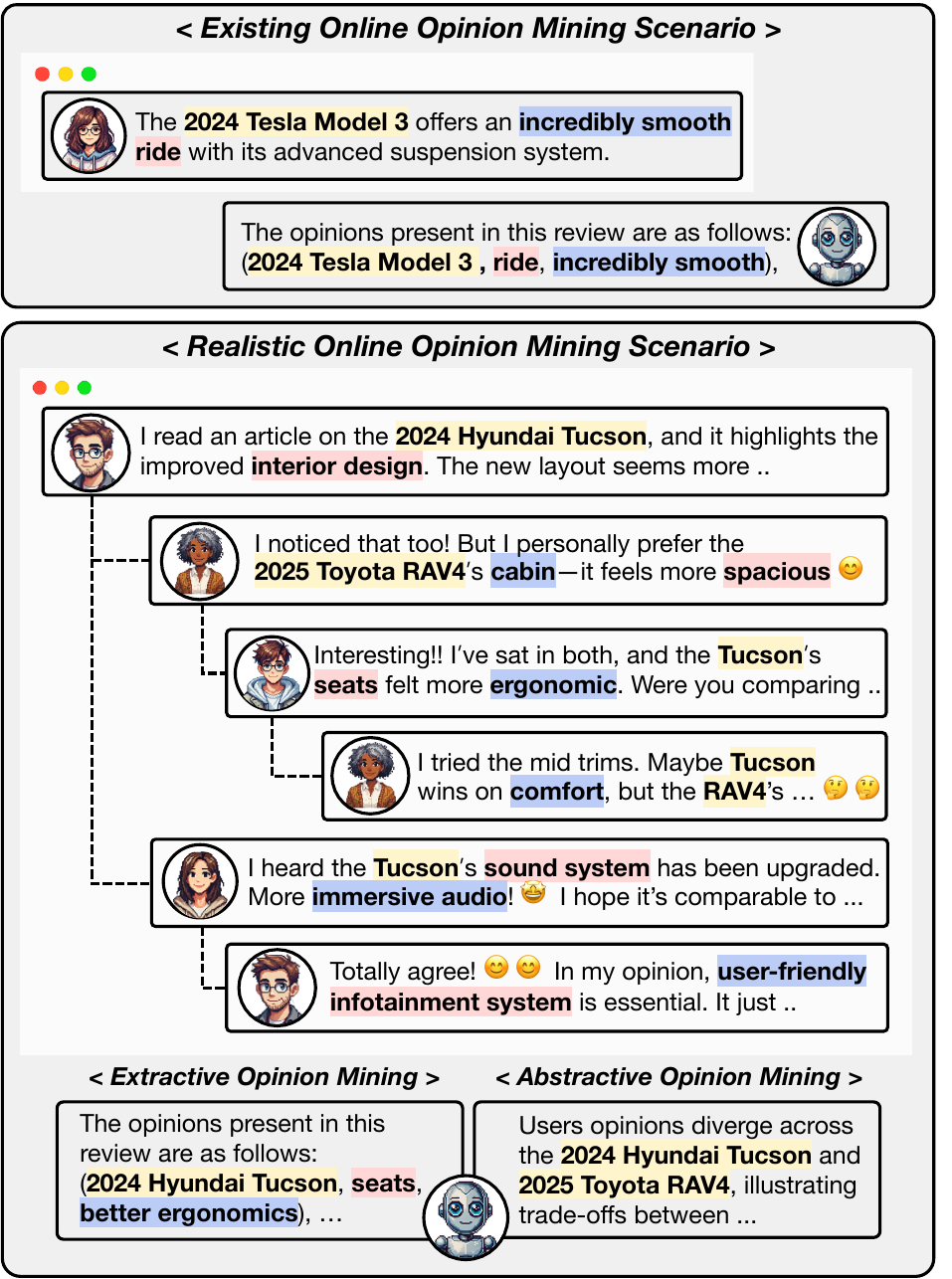}
    \caption{Existing opinion mining scenarios assume a simple input structure (\textbf{Upper}). In contrast, our study facilitates both extractive and abstractive opinion mining in complex, multi-threaded web discussions, enabling flexible and context-aware mining (\textbf{Lower}).}
    \label{fig:intro}
\end{figure}

The explosive growth of user-generated content has fundamentally transformed marketing strategies and business decision-making.
Companies now analyze vast amounts of user opinions scattered across platforms such as social media, review sites, and online communities to understand how consumers truly perceive their products and services \cite{rahayu2021impact, chen2022impact}.
As a result, \textit{opinion mining}—the task of extracting and analyzing opinions from online text—has become a core capability in today’s data-driven landscape.

Existing opinion mining approaches have primarily focused on identifying and extracting opinion expressions or spans within text \cite{irsoy-cardie-2014-opinion, xia2021unified, li2022diaasq, zhang2022survey, zhang-etal-2022-identifying}.
Over time, these methods have evolved to incorporate sentiment analysis \cite{zhao-etal-2020-spanmlt, zhang2021aspect, seo-etal-2024-make}, allowing for a deeper understanding of user preferences.

Despite these advances, existing approaches still face two critical limitations.
\textbf{(1) Underrepresentation of real-world input complexity}:
Previous benchmarks predominantly focus on single-sentence reviews \cite{Peng2019KnowingWH, cai2021aspect, zhang2021aspect} or preprocessed dialogue scenarios \cite{li2022diaasq}.
However, in real-world online environments, user opinions appear in far more complex and structurally diverse formats. 
In practice, opinion streams span multi-party threaded discussions, long-form narratives with interleaved pros/cons, and domain‐specific markers (e.g., emojis, slang, abbreviations) that introduce implicit sentiment signals (Figure~\ref{fig:intro} Lower).
The absence of a setting that comprehensively captures these realistic and diverse forms of opinion expression makes \textbf{it difficult to assess under what conditions and to what extent large language models (LLMs) can effectively perform opinion mining.}
This gap poses a significant challenge to evaluating the utility of LLMs and understanding their applicability to real-world applications.


\textbf{(2) Confinement to extraction-centric tasks}:
As mentioned earlier, most prior tasks have focused on extracting opinion spans or structured tuples from input texts. 
However, this extraction-centric approach can excessively simplify or compress the nuanced contextual information and emotional nuances that are essential for strategic decision-making.
For instance, the tuple (\textit{``Tesla Model 3'', ``interior'', ``larger than the previous model''}) fails to capture critical contextual background, such as whether the user inspected the vehicle in person or harbored an implicit purchase intent.
In real-world industry settings, marketers and product teams are more interested in cohesive, topic-level insights rather than isolated fragments of information  \cite{yuan2015users, santos2021consumer, han2023attribute}. 
These observations highlight the need to explore opinion mining paradigms that \textbf{move beyond raw extraction and aim to preserve the emotions, contextual subtleties, and user intent embedded} in real-world discourse.

To address these challenges, we propose \underline{\textbf{O}}nline \underline{\textbf{O}}pinion \underline{\textbf{M}}ining \underline{\textbf{B}}enchmark, named \textbf{\proposed}, a novel benchmark specifically tailored to evaluate the opinion mining capabilities of LLMs across realistic, complex, and diverse online scenarios. 
Unlike previous datasets, \proposed incorporates content from structurally distinct platforms, including blogs, review sites, Reddit threads, and YouTube comments, capturing long-form content as well as single \& multi-user interactions representative of authentic online discussions.
Each content instance is enriched with dual-layer annotations: (1) a structured set of (\textit{entity, feature, opinion}) tuples reflecting explicit user perspectives, and (2) a context-rich, opinion-centric \textit{insight} organized around key thematic topics from a marketer’s viewpoint.

Building upon this benchmark, we introduce two complementary tasks:
\textbf{(1) \subone (FOE)} evaluates whether LLMs can accurately extract structured opinions from online content and \textbf{(2) \subtwo (OIG)} assesses whether LLMs can mine high-level topics and insights from user opinions expressed in online content.
We conduct extensive experiments on ten proprietary and open-source LLMs to provide an in-depth analysis of their respective capabilities and limitations.
The evaluation results demonstrate that while the models struggle with  extracting structured opinions from online content, they exhibit relatively strong adaptability in synthesizing diverse user opinions into meaningful insights.
Based on these findings, we discuss key takeaways and potential future directions to further advance the field of opinion mining.
Specifically, our contributions are as follows:
\begin{itemize}[leftmargin=*,topsep=2pt,itemsep=2pt,parsep=0pt]
    \item We present \proposed, a realistic and richly annotated benchmark that evaluates LLMs across structurally diverse online content using both structured tuples and opinion-centric insights.
    
    \item We define two complementary tasks FOE and OIG—to jointly assess extraction and abstraction capabilities of LLMs from diverse online content.
    
    \item We extensively evaluate both proprietary and open-source LLMs, highlighting their strengths, limitations, and opportunities for further work.
\end{itemize}

\begin{figure*}[t]
    \centering
    \includegraphics[width=\textwidth]{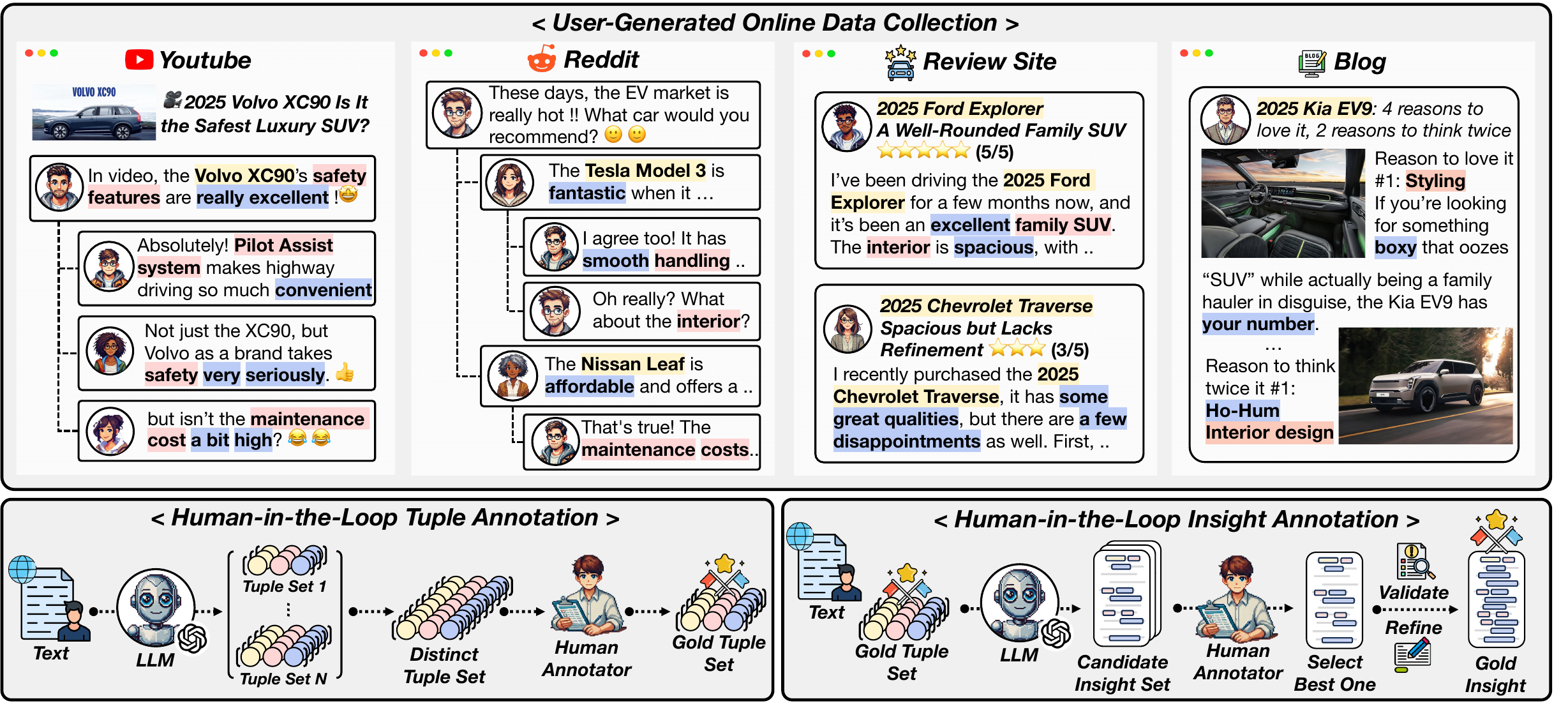}
    \caption{The overview of our \proposed benchmark construction pipeline.}
    \label{fig:oomb_overview}
\end{figure*}

%% file: TAB/020data_comparison.tex
\definecolor{darkgreen}{HTML}{006400}
\begin{table*}[!t]
\centering

\resizebox{\textwidth}{!}{%
\begin{tabular}{cccccccc}
\toprule
\multirow{2}{*}{\textbf{Benchmark}} & {\textbf{\#Test}} & {\textbf{Avg}} & {\textbf{Avg}} & {\textbf{Tuple}} & \multirow{2}{*}{\textbf{Content Types}} & \multicolumn{2}{c}{\textbf{Task}} \\
& \textbf{Examples} & \textbf{\#Tokens} & \textbf{\#Tuples} & \textbf{Components} & 
 & \textbf{Ext.} & \textbf{Abs.} \\
\midrule
\makecell{ASTE~\cite{Peng2019KnowingWH}}       & 1,468 & 15.7   & 1.7  & (\textit{a, o, s})      & Reviews  & \color{darkgreen} \ding{51} & \color{red} \ding{55} \\
\makecell{ACOS~\cite{cai2021aspect}}       & 1,399 & 15.2   & 1.5  & (\textit{a, c, o, s})   &Reviews & \color{darkgreen} \ding{51} & \color{red} \ding{55} \\
\makecell{ASQP~\cite{zhang2021aspect}}      & 1,081 & 14.9   & 1.5  & (\textit{a, c, o, s})   & Reviews  & \color{darkgreen} \ding{51} & \color{red} \ding{55} \\
\makecell{DiaASQ-EN~\cite{li2022diaasq}}  & 100   & 179.7  & 8.5  & (\textit{t, a, o, s})   & Conversation & \color{darkgreen} \ding{51} & \color{red} \ding{55} \\
\midrule
\makecell{\textbf{\proposed} (Ours)}   & 600   & 648.7  & 14.4 & (\textit{e, f, o})      & Reviews, Blogs, Conversation & \color{darkgreen} \ding{51} & \color{darkgreen}  \ding{51} \\
\bottomrule
\end{tabular}%
}
\caption{A comparison of our benchmark to existing opinion related benchmarks. 
Each tuple component represents the following: \textit{a}: aspect, \textit{c}: aspect category, \textit{o}: opinion, \textit{s}: sentiment, \textit{t}: target, \textit{e}: entity, and \textit{f}: feature.}
\label{tab:comparison_benchmark}
\end{table*}

%% file: 020dataset.tex
In this section, we introduce the construction of \textbf{\proposed}, a benchmark designed to effectively represent real-world online content.
Figure~\ref{fig:oomb_overview} illustrates the overall construction pipeline. 

\subsection{Data Collection}
\label{subsec:data_col}
To reflect realistic user-generated content and a wide range of online structures, we collect textual data from four different sources: \textit{Blog}, \textit{Review Site}, \textit{Reddit}, and \textit{YouTube}. 
Blog and review site provide detailed long-form posts and specific car reviews, while reddit and youtube capture multi-threaded and single-threaded discussions, respectively.
Specifically, we curate the sources of each website from Feedspot,\footnote{https://www.feedspot.com/} a platform that organizes and manages content across various topics. 
More details about our data collection process and sources are provided in Appendix \ref{subsec:data_source_details}.

\subsection{Data Annotation}
\label{subsec:data_anno}
For each collected user-generated content, we construct a dual-layer annotation for each content, consisting of both structured opinion tuples and free-form opinion-centric insights.
Following recent studies demonstrating that LLMs with advanced reasoning capabilities can serve as effective tools for data annotation \cite{he-etal-2024-annollm, tan2024largelanguagemodelsdata}, we adopt a human-in-the-loop process in which LLM is first used as the initial annotator, and human annotators verify and refine them to ensure high-quality, reliable labels.
The detailed annotation process is described in Appendix~\ref{subsec:appendix_annotation_refinement}.

\paragraph{Entity-feature-opinion tuple}
We annotate each content with structured \textit{(entity, feature, opinion)} tuples that capture user perspectives.
In contrast to the commonly used (aspect, category, opinion, sentiment) schema, this design more closely reflects how real users express opinions. 
Typically, they do so without explicit category or sentiment labels but rather through direct mentions of entity features.
Specifically, \textit{entity} refers to the specific subject or object under discussion in the content (e.g., ``Volvo XC90'').
\textit{feature} indicates a characteristic, attribute, or component of the entity that a user mentions or evaluates (e.g., ``interior design''). 
\textit{opinion} represents the subjective or objective judgment, reaction, experience, evaluation, or feedback regarding a feature (e.g., ``luxurious''). 
If a feature is implicit and does not appear explicitly in the text, it is labeled as ``NULL'' following \cite{cai2021aspect}.
In all other scenarios, each component of the tuple is assumed to be an explicitly mentioned span in the content.

\paragraph{Tuple annotation}
To maximize the coverage and diversity of (entity, feature, opinion) tuples from each input content, we perform five rounds of zero-shot prompting using \gptmini\footnote{\texttt{gpt-4o-mini-2024-07-18}}.
We then take the union of all generated tuples and remove duplicates to form a distinct preliminary set of tuples.
Subsequently, five trained human annotators review every candidate tuple for correctness, eliminate hallucinated entries, and complement any missing tuples.
To support consistent decision-making, we design detailed task-specific annotation guidelines and conduct a one-week training session for all annotators, including case-based instruction and edge-case discussions.
This process was applied to every content, thereby ensuring high coverage, consistency, and reliability in the final annotations.

\paragraph{Opinion-centric insight}
We annotate each content with an opinion-centric insight, a free-form text that organizes diverse opinions into high-level topics for meaningful insights.
Specifically, from a marketing manager’s perspective, opinions are grouped into broad categories, highlighting frequently mentioned or standout aspects to reveal key trends. 
This insight follows a \textit{three-to-five-line form}, providing a cohesive structure for clear and concise representation of core discussions.

\paragraph{Insight annotation}
We generate five independent candidate insights using the input content and the associated final set of tuples as input.
Then, the same five human annotators review each insight from the perspective of a marketing manager and select the highest-quality one that best captures the opinions in the final tuples at the topic level. 
Similar to the tuple annotation process, we design detailed annotation guidelines to ensure consistent decision-making, and all five annotators undergo a one-week training session. 
If all candidate insights are deemed insufficient in quality, the annotators collaboratively rewrite a new insight that more accurately reflects the key insights.
For the selected insight, the annotators collaboratively refine and finalize it by checking for missing opinions, eliminating hallucinations, and ensuring conciseness in a three-to-five-line format.
\input{TAB/021iaa}
\subsection{Inter-Annotator Agreement Analysis}
\label{subsec:human_verifi}
To further ensure the reliability of our annotation process, we conduct an inter-annotator agreement study. 
Five trained annotators participated in this analysis across 100 randomly sampled contents. 
We assess agreement at three different stages of the annotation process: (1) Span-level tuple agreement, (2) Preferred insight agreement, and (3) Insight semantic consistency.
Table~\ref{tab:iaa_results} demonstrate that annotators reached substantial agreement in both structured tuple extraction and insight selection, while finalized insights exhibited strong semantic consistency across annotators. 
This comprehensive analysis confirms that our benchmark was constructed through a highly reliable process.

\subsection{Statistics and Analysis}
\label{subsec:data_stat_anal}
As shown in Table \ref{tab:comparison_benchmark}, unlike previous benchmarks, \proposed features substantially longer average token lengths and a significantly higher number of tuples, making it considerably more challenging. 
Additionally, it covers a broader and more diverse range of content types while supporting two tasks: extraction and abstraction.
This dual-task setup enables the evaluation of LLMs in more realistic settings by reflecting the complexity and variability of real-world opinion expressions.
Detailed our benchmark statistics are presented in Appendix~\ref{subsec:dataset_statistics}.

%% file: TAB/021iaa.tex
\begin{table}[t]
\centering
\small
\begin{tabular}{lcc}
\toprule
\textbf{Task} & \textbf{Metric} & \textbf{Score} \\
\midrule
Span-level Tuple Agreement & Fleiss' Kappa & 0.6821 \\
Preferred Insight Agreement  & Fleiss' Kappa & 0.7695   \\
Insight Semantic Consistency & BERTScore & 0.9297 \\
\bottomrule
\end{tabular}
\caption{Inter-annotator agreement score across three different annotation stages (\textit{p-value} < 0.05). Detailed description of the analysis is provided in the Appendix~\ref{subsec:appendix_annotation_refinement}}
\label{tab:iaa_results}
\end{table}


%% file: 030task_eval.tex
\input{TAB/051extractive_main}

\subsection{\subone (FOE)}
\label{subsec:foe}
\paragraph{Problem formulation}
Similar to existing opinion mining approaches \cite{fan-etal-2019-target, xia2021unified}, this task aims to enable LLMs to accurately identify and extract structured set of opinion tuples from the given input content.
Formally, given a content \(c\), our goal is to identify and extract a set of tuples \(\mathcal{T} = \{(e_i, f_i, o_i)\}_{i=1}^{N}\), where \(e_i\) represents the entity, \(f_i\) the feature, and \(o_i\) the opinion.

\paragraph{Evaluation metrics}
To evaluate structured opinion extraction capabilities of LLMs, we utilize three types of tuple matching evaluation methods.
\textbf{(1) Exact Match (EM)}: Consistent with existing opinion-related extraction tasks \cite{zhang2022survey, xia2021unified}, a predicted tuple is considered correct only if all its elements exactly match the corresponding elements in the gold tuple.
\textbf{(2) Relaxed Match (RM)}: 
To provide a more flexible evaluation beyond strict exact matching, we evaluate the similarity of each tuple component using both lexical and semantic matching.
A tuple is considered a relaxed match if the similarity score of all its components exceeds a predefined threshold of 0.7, formally defined as:
\begin{equation*}
\small
\text{RM}(t_p, t_g) =
\begin{cases} 
1, & \text{if } \forall x \in \{e, f, o\}, \; \text{Sim}(x_p, x_g) \geq 0.7 \\
0, & \text{otherwise}
\end{cases}
\end{equation*}
where \( t_p = (e_p, f_p, o_p) \) and \( t_g = (e_g, f_g, o_g) \) are the predicted and gold tuples, respectively.
Drawing on recent works \cite{han2023information, li2024simple}, we utilize the Python's difflib library\footnote{https://docs.python.org/3/library/difflib.html} to compute token-level overlap scores for lexical similarity (L-RM), while employing a Sentence Transformer\footnote{\texttt{sentence-transformers/all-MiniLM-L6-v2}} for semantic similarity (S-RM).
\textbf{(3) Contextual Match (CM)}: 
Inspired by \cite{fu2023gptscoreevaluatedesire, fane-etal-2025-bemeae}, we design a method that leverages the reasoning capabilities of LLMs to match tuples in a manner similar to human judgment. 
Specifically, we utilize \gptfouro\footnote{\texttt{gpt-4o-2024-08-06}} to evaluate both predicted and gold tuples, enabling the model to count how many tuples match.
This metric allows recognition of semantically equivalent tuples even when surface forms differ significantly, used the prompt shown in Table~\ref{tab:gm_prompt}.
Note that for both RM and CM, we measure recall by counting each gold tuple at most once to avoid double counting, even if multiple predicted tuples match the same gold tuple.
For all evaluation metrics, we primarily use the F1 score while also reporting precision and recall.

\subsection{\subtwo (OIG)}
\label{subsec:cos}
\paragraph{Problem formulation}
This task aims to analyze whether LLMs can group scattered opinions from user-generated online content into high-level topics, providing context-aware and meaningful insights.
Formally, given content \(c\), our objective is to generate a insightful text \(I\) that cohesively encapsulates user opinions into high-level topics.

\paragraph{Evaluation metrics}
To broadly assess the quality of opinion-centric insights generated by the model across various aspects, we employ both lexical and semantic automated evaluation metrics.
For lexical evaluation, we adopt \textbf{ROUGE-1, 10, L} \cite{lin-2004-rouge}, which measure word overlap between the reference and generated insights.
For semantic evaluation, we leverage \textbf{BERTScore (BS)} \cite{Zhang2019BERTScoreET} and \textbf{A3CU} \cite{liu-etal-2023-towards-interpretable}.
BS computes similarity between the reference and generated texts using contextual embeddings, while A3CU compares texts without extracting atomic content units, providing a human-aligned assessment of content similarity.
For both ROUGE and A3CU, we report F1 scores.
Moreover, to ensure a systematic and comprehensive evaluation, we also conduct reference-free assessments using an LLM-as-a judge.
Inspired by~\cite{siledar2024one}, we design the following six well-defined criteria: \textit{Faithfulness, Coverage, Specificity, Insightfulness, Intent} and \textit{Fluency}.
This analysis extends beyond automated lexical and semantic metrics, providing a broader perspective on the abstractive opinion mining capabilities of LLMs.
A detailed description is provided in Appendix~\ref{para:LLM_judge_eval_criteria}.

\input{TAB/053correlation}

\subsection{Experimental setup}
\paragraph{Models}
We conduct extensive experiments on two types of LLMs: \textbf{(1) Proprietary LLMs} that are available via APIs, such as GPT-4o-mini, GPT-4o \cite{openai2024gpt4technicalreport}, and Claude 3.5 Haiku, Sonnet \cite{anthropic2024}. 
\textbf{(2) Open-source LLMs} such as Llama-3-Instruct (8B, 70B, \citealt{grattafiori2024llama3herdmodels}), Gemma 2-it (9B, 27B, \citealt{gemmateam2024gemma2improvingopen}), Qwen2.5-7B-Instruct \cite{yang2024qwen2technicalreport}, and DeepSeek-7B-chat \cite{bi2024deepseek}.

\paragraph{Implementation details}
Following recent studies demonstrating the reasoning capabilities of LLMs in zero-shot settings \cite{wang2024user, qinsysbench, liu-etal-2024-unleashing-power, chhabra-etal-2024-revisiting}, we perform both tasks using zero-shot prompting.
This means the models rely solely on their pre-trained knowledge without any task-specific fine-tuning.
To ensure consistent and reliable performance across all experiments, we set the temperature to 0 for all generations.
Our more detailed experimental setup presented in Appendix \ref{sec:experimental_details}.

%% file: TAB/051extractive_main.tex
\begin{table*}[!t]
\centering
\small
\resizebox{0.99\linewidth}{!}{
\begin{tabular}{lcccccccccccc}
\toprule
\multirow{2.5}{*}{\textbf{Models}}
& \multicolumn{3}{c}{\textbf{EM}} 
& \multicolumn{3}{c}{\textbf{L-RM}} 
& \multicolumn{3}{c}{\textbf{S-RM}} 
& \multicolumn{3}{c}{\textbf{CM}} \\
\cmidrule(lr){2-4}\cmidrule(lr){5-7}\cmidrule(lr){8-10}\cmidrule(lr){11-13}
& \textbf{Pre} & \textbf{Rec} & \textbf{F1} 
& \textbf{Pre} & \textbf{Rec} & \textbf{F1} 
& \textbf{Pre} & \textbf{Rec} & \textbf{F1} 
& \textbf{Pre} & \textbf{Rec} & \textbf{F1} \\ \midrule
\multicolumn{13}{c}{\textbf{\textit{Proprietary LLMs}}} \\ \midrule
\gptmini & 3.91 & 1.99 & 2.62 & 11.29 & 5.70 & 7.52 & 15.50 & 7.76 & 10.27 & \textbf{65.07} & 36.61 & 43.19 \\  
\gptfouro & 7.27 & \underline{5.18} & \underline{6.02} & \underline{15.86} & \underline{11.39} & \underline{13.20} & \underline{21.23} & \underline{15.31} & \underline{17.71} & 59.34 & \textbf{45.88} & \textbf{48.28} \\   
\claudehaiku & \underline{6.13} & 3.02 & 4.01 & 15.02 & 7.51 & 9.94 & 20.60 & 10.36 & 13.68 & \underline{63.70} & 37.25 & 43.62 \\  
\claudesonnet & \textbf{11.12} & \textbf{6.32} & \textbf{7.97} & \textbf{22.97} & \textbf{13.02} & \textbf{16.46} & \textbf{29.30} & \textbf{16.52} & \textbf{20.93} & 62.00 & \underline{39.83} & \underline{44.90} \\   \midrule
\multicolumn{13}{c}{\textbf{\textit{Open-source LLMs}}} \\ \midrule
\llamaeight & \textbf{8.49} & \textbf{6.28} & \textbf{7.17} & \textbf{16.75} & \textbf{12.18} & \textbf{14.02} & \textbf{21.33} & \underline{15.43} & \textbf{17.80} & 51.92 & 42.51 & 43.18 \\  
\llamaseventy & \underline{7.26} & \underline{5.57} & \underline{6.27} & \underline{15.21} & \underline{11.66} & \underline{13.13} & \underline{19.94} & 15.18 & 17.15 & \underline{53.15} & \underline{42.55} & \textbf{43.67} \\ 
\gemmanine & 6.37 & 4.51 & 5.25 & 14.17 & 10.17 & 11.78 & 17.73 & 12.59 & 14.64 & \textbf{53.71} & 41.93 & \underline{43.61} \\  
\gemmatwentyseven & 7.05 & 5.61 & 6.20 & 14.33 & 11.77 & 12.82 & 19.29 & \textbf{15.83} & \underline{17.25} & 52.58 & \textbf{42.98} & 43.42 \\  
\qwen & 6.55 & 4.18 & 5.05 & 12.97 & 8.32 & 10.02 & 18.22 & 11.83 & 14.18 & 52.48 & 39.15 & 41.14 \\  
\deepseek & 3.00 & 1.63 & 2.07 & 5.86 & 3.13 & 4.02 & 8.20 & 4.34 & 5.61 & 49.25 & 30.33 & 33.12 \\  
\bottomrule
\end{tabular}
}
\caption{Performance comparison of various LLMs for the FOE task across diverse tuple matching metrics.}
\label{tab:extractive_main}
\end{table*}

%% file: TAB/053correlation.tex
\begin{table}[!t]
\centering
\small
\begin{tabular}{lccc}
\toprule
\textbf{Metric} & \textbf{Pearson \(r\)} & \textbf{Spearman \(\rho\)} & \textbf{Kendall \(\tau\)} \\
\midrule
\textbf{EM} & 0.4505 & 0.4722 & 0.4215 \\
\textbf{L-RM} & 0.4584 & 0.4754 & 0.4244   \\
\textbf{S-RM} & \underline{0.5514} & \underline{0.5531} & \underline{0.4937} \\
\textbf{CM} & \textbf{0.8337} & \textbf{0.8155} & \textbf{0.7279} \\
\bottomrule
\end{tabular}
\caption{
Correlation coefficients between each metric and human judgment (\textit{p-value} < 0.05) based on pairwise comparisons by five human evaluators.
Detailed experimental settings are provided in Appendix \ref{para:correlation}.}
\label{tab:correlation_results}
\end{table}

%% file: 050results.tex
\begin{figure}[t]
\centering
\includegraphics[width=0.95\linewidth]{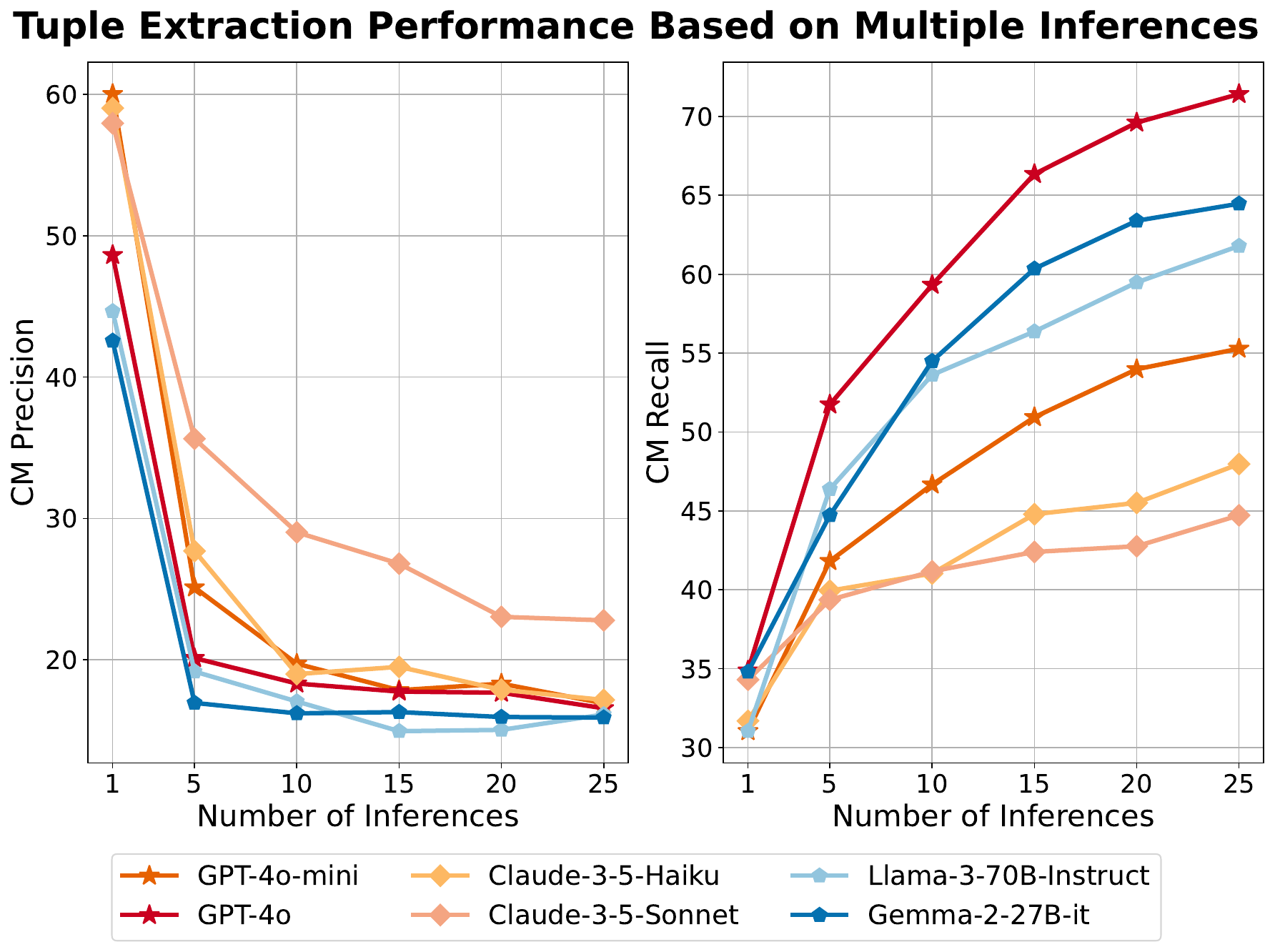}
\caption{Performance comparison of various LLMs on the FOE task, increasing the numbers of inferences.}
\label{fig:multi_inference}
\end{figure}

In this section, we present the main findings of our study. 
Each subsection addresses the research question \textit{“Can LLMs serve as effective online opinion miners?”} from various perspectives, supported by detailed experimental results and analyses.

\subsection{RQ1: What makes it challenging for LLMs to extract structured opinions?}
\paragraph{Performance on tuple extraction}
As shown in Table \ref{tab:extractive_main}, LLMs consistently struggle to extract structured opinions. 
Specifically, even the best-performing model fails to achieve an F1 of 30 on both the rigid EM metric and the more relaxed RM metric, demonstrating significantly low performance.
In contrast, employing the CM leads to a notable and consistent improvement in both tuple matching accuracy and overall recall.
This metric effectively leverages LLMs' reasoning capabilities to mirror human judgment and has been shown to align most closely with human evaluations (see Table \ref{tab:correlation_results}).
Nevertheless, even with CM, \textbf{most models fail to both accurately predict the correct tuples and comprehensively cover all tuples present in the input content}, revealing inherent limitations in LLMs' extraction capabilities.
This highlights that structured opinion extraction remains a highly complex and challenging task for LLMs, particularly in the context of realistic online content.

\paragraph{Effect of multiple inference on tuple extraction performance}
To investigate how extensively an LLM can extract structured tuples from content, we perform multiple inference iterations per single input and measure the model's extraction performance.
For evaluation, we take the union of all tuples generated across iterations, remove duplicates, and consider only the unique set of \textit{(entity, feature, opinion)} tuples.
To capture a broader range of tuples, we set the temperature to 1.0 during inference.
As shown in Figure \ref{fig:multi_inference}, most models generate a significantly larger number of predicted tuples as the number of inference iterations increases, but the number of correctly matched tuples does not keep pace.
Notably, recall improves significantly across most models but eventually reaches a plateau, where the rate of increase diminishes. 
\textbf{This implies that LLMs recognize a fixed set of opinions within the content, making it challenging to cover every opinion merely by increasing the number of inference iterations.}
Therefore, improving the extraction capabilities of LLMs requires exploring alternative strategies beyond merely repeating the inference process.

\input{TAB/054casestudy}
\paragraph{Case study: LLMs' extraction capability}
We conduct a case study to identify key failure patterns that limit LLMs' ability to extract structured opinions. 
{Table~\ref{tab:casestudy} illustrates} the comparison between the gold tuples and GPT-4o's predicted tuples for actual input content across EM, L-RM, S-RM, and CM. 
Despite being explicitly instructed in the input prompt to extract spans as-is, LLM often produces semantically related but non-identical spans—``interior materials'' instead of “interior”—substitutes related concepts such as ``engine'' for ``power'', and even hallucinates opinions like ``excellent'' in place of ``reclaims a physical volume knob''.
These patterns indicate that LLMs tend to transform or reinterpret textual information rather than extracting it verbatim as structured tuples.
Such behavior underscores a fundamental limitation of LLMs in this task and suggests that structured extraction may not be an effective approach for opinion mining with LLMs.

\input{TAB/052abstractive_main}

\subsection{RQ2: How effectively can LLMs generate opinion-centric insight?}
\paragraph{Automated evaluation results}
Table \ref{tab:abstractive_main} reports the performance for the OIG task, using both lexical and semantic evaluation metrics. 
While models show strong word-level overlap (R-1 and R-L) in their insights, they exhibit significantly lower bigram recall (R-2), highlighting difficulty in sustaining coherent phrase structures.
Additionally, they achieve relatively high BS; their performance on A3CU remains substantially lower, suggesting that LLMs often capture surface-level semantic similarity but struggle to reflect deeper, human-aligned content understanding. 
Thus, to thoroughly gauge LLMs’ abstractive strengths, particularly in capturing intent, subtle sentiment shifts, and deeper insights beyond surface semantics, a multifaceted evaluation framework is needed.

\begin{figure}[t]
\centering
\includegraphics[width=0.99\linewidth]{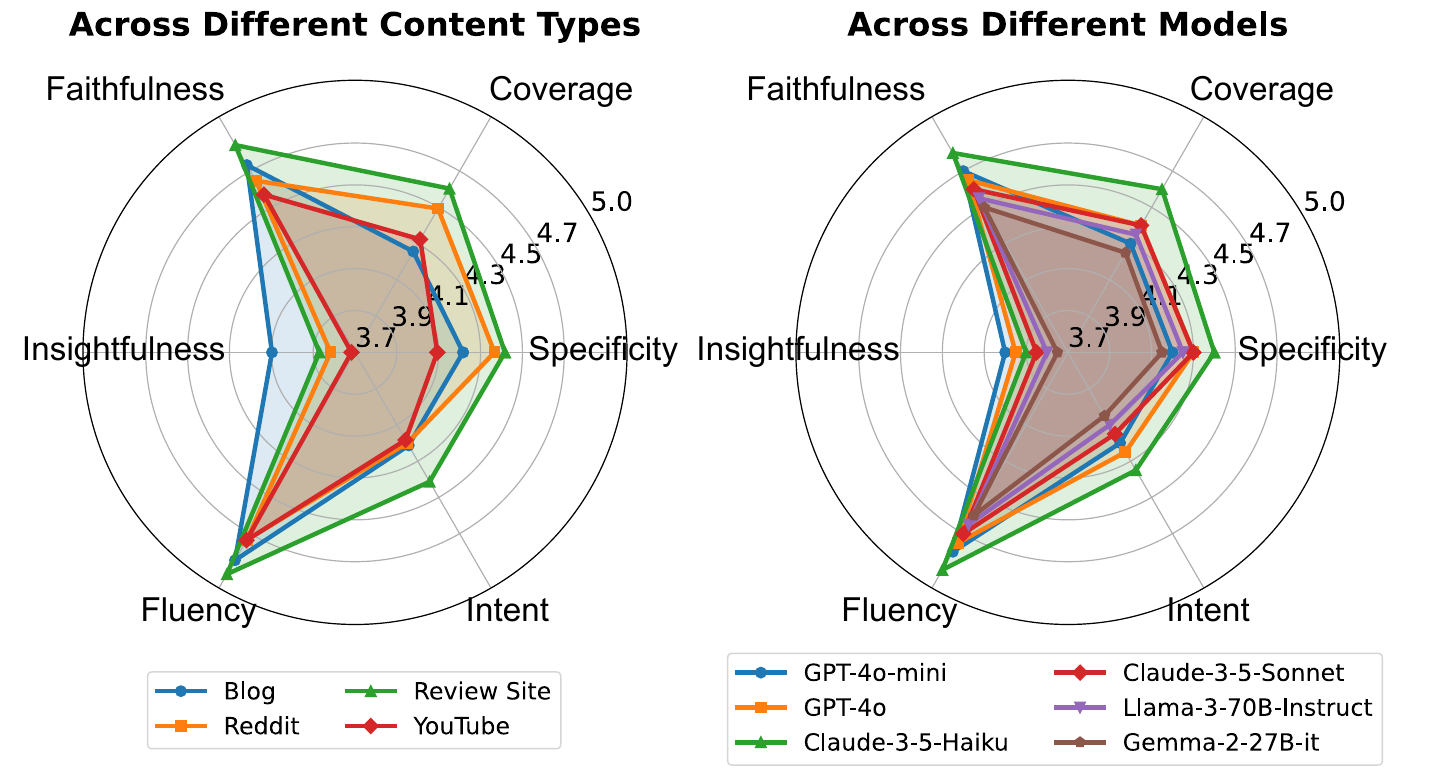}
\caption{Radar charts for LLM-as-a-judge evaluations of the OIG task. Comparison of the average model performance across different content types \textbf{(Left)}. Comparison of performance across different models \textbf{(Right)}.
}
\label{fig:llm_judge_radar}
\end{figure}
\paragraph{LLM-as-a-judge evaluation across multiple perspectives}
To comprehensively analyze how well models generate abstractive opinion-centric insights, we conduct a reference-free evaluation using an LLM as the judgemeter.
From the results in Figure \ref{fig:llm_judge_radar}, we derive the following key conclusions:
\textbf{(1) LLMs consistently provide natural and readable insights while preserving the original content without distortion or unnecessary modification.}
This demonstrates their strength in faithfulness and fluency, ensuring that the generated insights remain accurate and coherent.
\textbf{(2) However, LLMs struggle to capture implicit user intentions, nuanced expressions, and meaningful insights that are not explicitly stated in the input content.}
This limitation is reflected in lower scores for insightfulness and intent, indicating that while LLMs can summarize well, they lack deeper abstraction and contextual understanding.

\begin{figure*}[t]
\centering
\includegraphics[width=\textwidth]{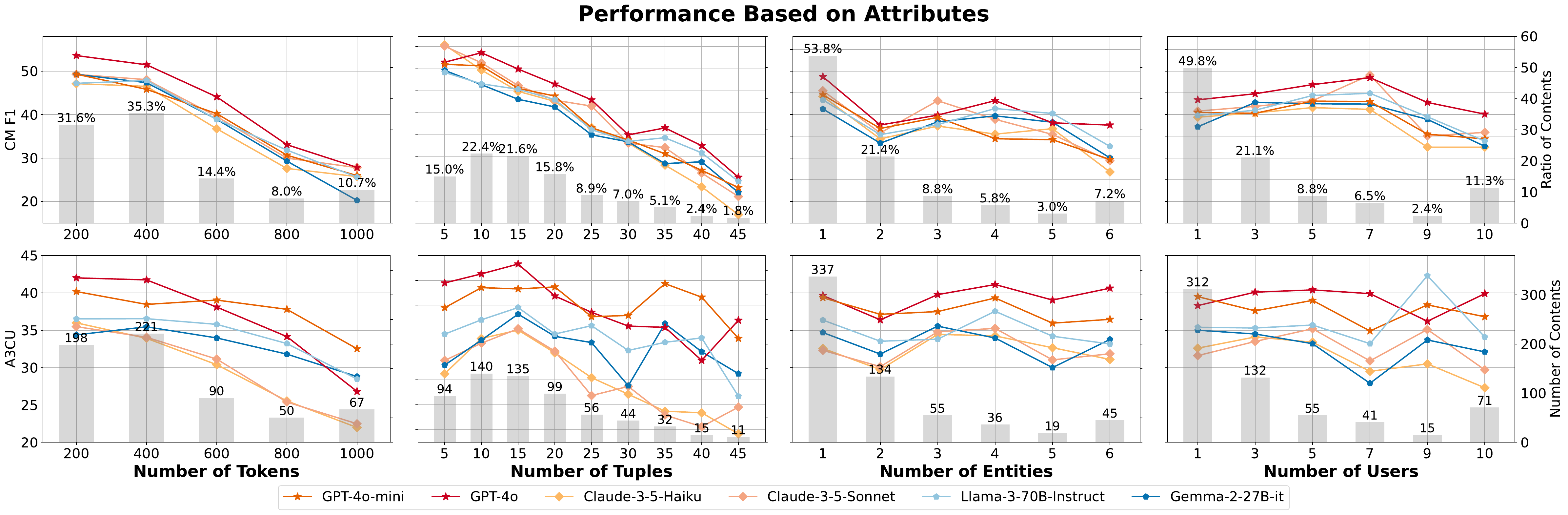}
\caption{Performance comparison of various LLMs based on changes in different attributes within online content.
CM F1 scores for the FOE task \textbf{(Upper)}, and A3CU scores for the OIG task \textbf{(Lower)}.}
\label{fig:attribute_performanc}
\end{figure*}

\begin{figure}[h]
\centering
\includegraphics[width=0.99\linewidth]{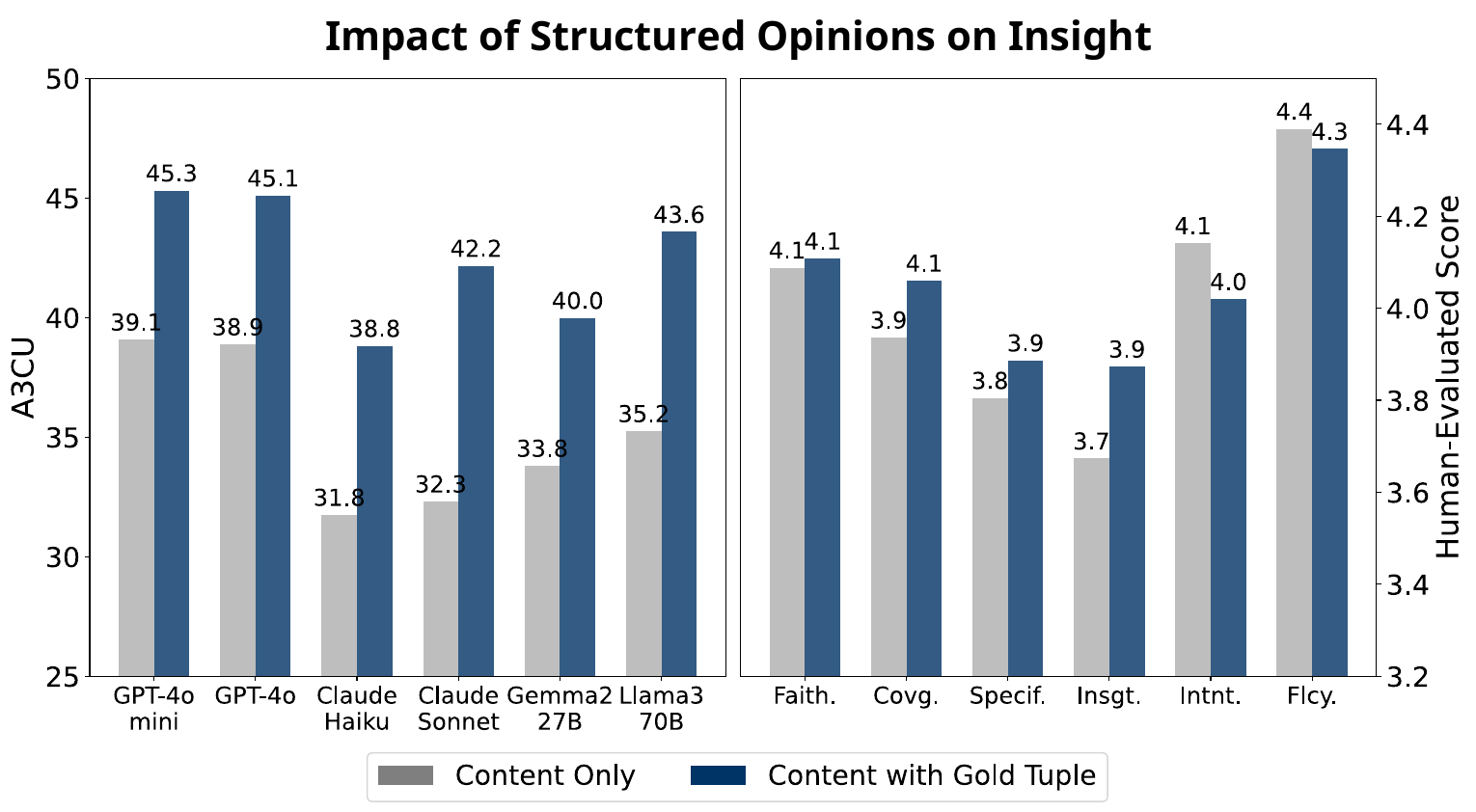}
\caption{Comparison of the OIG task performance between content-only input and input with gold tuples. 
Automated metric performance by model \textbf{(Left)} and human evaluation of \gptfouro outputs \textbf{(Right)}.
Detailed human evaluation is presented in Appendix \ref{para:amazon_hman_eval}.}
\label{fig:a3cu_comparison}
\end{figure}

\paragraph{Impact of structured opinions on insight generation}
Figure~\ref{fig:a3cu_comparison} demonstrates that augmenting opinion tuples during opinion-centric insight generation not only leads to substantial improvements in automatic evaluation metrics, but also provides practical benefits from a user perspective.
In particular, the notable gains in \textit{coverage} and \textit{insightfulness} suggest that the model becomes more effective at capturing key opinions and delivering more informative opinion-centric insights.
Conversely, slight decreases in \textit{intent} and \textit{fluency} indicate that the added structure may sometimes interfere with natural expression and tone preservation.
These results suggest that integrating structured opinion tuples into the insight generation pipeline is a key strategy for effective opinion mining, while also \textbf{highlighting the need for continued research into both the extractive and abstractive capabilities of LLMs.}

\subsection{RQ3: Do LLMs effectively adapt to diverse online text environments?}
To assess the adaptability of LLMs to the highly varied nature of online content, we analyze their performance across several dimensions.
Figure \ref{fig:attribute_performanc} presents a comparative analysis of how different LLMs perform when these attributes vary.

\paragraph{LLMs struggle with dense and lengthy content}
Scenarios involving long content or a large number of tuples inherently present verbose and opinion-rich online content. 
The more densely packed the information within these texts, the more LLMs noticeably struggle to extract opinions and derive insights.
This observation highlights the considerable challenges LLMs face in opinion mining when dealing with highly condensed and information-dense content. 
Therefore, it is crucial to explore more adaptive and effective mining approaches tailored to such complex scenarios (e.g., long-form user-generated texts, multi-thread discussions).

\paragraph{LLMs are robust in complex entity and multi-user environments}
In contrast, in environments where multiple users participate or the entity complexity increases, both extraction and opinion insight generation show a relatively weaker downward trend. 
These results indicate that in settings such as forum discussions or multi-user comment threads, LLMs do not face significant challenges in extracting information and synthesizing opinions. 
This finding implies that forums and communities tend to consist of simple comments or relatively easily recognizable subtopics, allowing LLMs to identify and summarize key opinions with ease.

%% file: TAB/054casestudy.tex
\begin{table}[t]
\centering
\resizebox{0.99\linewidth}{!}{
\begin{tabular}{C{2cm}C{2cm}C{2cm}C{2cm}} 
\toprule
\multicolumn{4}{p{10cm}}{\textbf{Content}: Interior materials remain perfectly adequate for the price of the truck and Honda’s reputation.} \\
\multicolumn{4}{p{10cm}}{\textbf{Gold}: (`2021 honda ridgeline', `interior', `adequate for the price')
}\\
\multicolumn{4}{p{10cm}}{\textbf{Predicted}: (`2021 honda ridgeline', `interior materials', `perfectly adequate')}\\
\cdashline{1-4}
EM: \color{red} \ding{55} & L-RM: \color{red} \ding{55} & S-RM: \color{darkgreen} \ding{51} & CM: \color{red} \ding{55} \\
\midrule

\multicolumn{4}{p{10cm}}
{\textbf{Content}: It’s not a tower of power by any stretch but gets the job done, even with a payload of swingset.} \\
\multicolumn{4}{p{10cm}}{\textbf{Gold}: (`2021 honda ridgeline', `power', `gets the job done')}\\
\multicolumn{4}{p{10cm}}{\textbf{Predicted}: (`2021 honda ridgeline', `engine', `gets the job done')}\\
\cdashline{1-4}
EM: \color{red} \ding{55} & L-RM: \color{red} \ding{55} & S-RM: \color{red} \ding{55} & CM: \color{darkgreen} \ding{51} \\
\midrule

\multicolumn{4}{p{10cm}}
{\textbf{Content}: The major update to the Ridgeline for the 2021 model year isn’t in its powertrain (remains the same), interior (reclaims a physical volume knob)} \\
\multicolumn{4}{p{10cm}}{\textbf{Gold}: (`2021 honda ridgeline', `interior', `reclaims a physical volume knob')}\\
\multicolumn{4}{p{10cm}}{\textbf{Predicted}: (`2021 honda ridgeline', `volume knob', `excellent')}\\
\cdashline{1-4}
EM: \color{red} \ding{55} & L-RM: \color{red} \ding{55} & S-RM: \color{red} \ding{55} & CM: \color{red} \ding{55} \\
\bottomrule
\end{tabular}
}
\caption{Examples of comparisons between gold and predicted tuples for structured opinion extraction.}
\label{tab:casestudy}
\end{table}

%% file: TAB/052abstractive_main.tex



\begin{table}[!t] 
\centering
\resizebox{0.90\linewidth}{!}{
\begin{tabular}{lPPPPP} 
\toprule
\multirow{2.5}{*}{\textbf{Models}}
& \multicolumn{3}{c}{\textbf{Lexical}} & \multicolumn{2}{c}{\textbf{Semantic}} \\ 
\cmidrule(lr){2-4} \cmidrule(lr){5-6}
& R-1 & R-2 & R-L & BS & {A3CU} \\ \midrule
\multicolumn{6}{c}{\textbf{\textit{Proprietary LLMs}}} \\ \midrule
\gptmini & \underline{39.30} & \underline{14.05} & \underline{34.58} & \textbf{90.35} & \textbf{38.49} \\
\gptfouro & \textbf{39.36} & \textbf{14.77} & \textbf{34.85} & \underline{89.86} & \underline{38.39} \\
\claudehaiku & 33.47 & 10.06 & 29.00 & 88.50 & 31.91 \\
\claudesonnet & 33.60 & \textcolor{white}{0}9.47 & 29.53 & 88.79 & 31.67 \\ \midrule
\multicolumn{6}{c}{\textbf{\textit{Open-source LLMs}}} \\ \midrule
\llamaeight & \underline{37.48} & \textbf{13.15} & \underline{33.43} & 89.91 & 30.50 \\ 
\llamaseventy & \textbf{37.61} & \underline{13.04} & \textbf{33.18} & \textbf{90.15} & \underline{31.48} \\
\gemmanine & 35.03 & 11.47 & 30.99 & 88.25 & 31.16 \\
\gemmatwentyseven & 35.40 & 11.69 & 31.02 & \underline{90.08} & \textbf{34.09} \\
\qwen & 33.84 & 10.87 & 27.94 & 89.56 & 25.34 \\
\deepseek & 35.03 & 10.68 & 30.72 & 76.89 & 25.80 \\
\bottomrule
\end{tabular}
}
\caption{Performance comparison of various LLMs for the OIG task across automated evaluation metrics.}
\label{tab:abstractive_main}
\end{table}

%% file: 060relatedwork.tex
\label{subsec:relatedwork}
Early studies on opinion mining \cite{pang2008opinion} primarily focused on identifying and classifying opinion-related expressions or spans within text \cite{yang-cardie-2013-joint, irsoy-cardie-2014-opinion, katiyar-cardie-2016-investigating, xia2021unified, liu2021comparative, zhang-etal-2022-identifying}. 
In particular, extracting opinions about specific aspects of products and services received significant attention \cite{fan-etal-2019-target, wu-etal-2020-deep, zhao-etal-2020-spanmlt, chen-etal-2020-synchronous}, and subsequent work extended this to jointly predict sentiment, enabling more complex and insightful analyses \cite{Peng2019KnowingWH, cai2021aspect, zhang2021aspect, li2022diaasq, kim-etal-2024-self-consistent, seo-etal-2024-make, bai-etal-2024-compound}.

Recently, large language models (LLMs) \cite{openai2024gpt4technicalreport, grattafiori2024llama3herdmodels, gemmateam2024gemmaopenmodelsbased} have demonstrated remarkable zero-shot and in-context learning capabilities across a range of tasks, including information extraction \cite{kim2024verifinerverificationaugmentednerknowledgegrounded, perot-etal-2024-lmdx, liu-etal-2024-unleashing-power} and abstractive summarization \cite{chhabra-etal-2024-revisiting, tang2024promptedaspectkeypoint, siledar2024one}. 
While these advances suggest that LLMs have great potential in opinion mining, existing benchmarks fall short of capturing the complexity of real-world inputs and remain focused on simplified, structured extraction settings. 
As a result, they fail to fully assess the true potential of LLMs in this domain.
To bridge this gap, we propose the \proposed benchmark, which encompasses a wide spectrum of realistic online content and enables comprehensive investigation of both extraction and abstractive capacities of LLMs.

%% file: 070conclusion.tex
In this paper, we introduce \proposed, a novel benchmark designed to assess LLMs’ capabilities in both structured opinion extraction and insight-oriented opinion generation across diverse and realistic online content scenarios.
To the best of our knowledge, \proposed\ is the first comprehensive benchmark for evaluating LLMs in both structured and abstractive opinion mining tasks under real-world conditions.
Our research reveals the dual challenge of precise opinion extraction and contextual insight generation, highlighting the need for future research to improve the effectiveness of both approaches.
This work lays the foundation for LLM-based opinion mining and serves as a stepping stone for future research in this field.

%% file: 080limitations.tex
Despite its contributions, this study has several limitations, each of which also suggests promising directions for future research and practical extensions.
First, although \proposed includes a diverse range of user-generated online content, it is currently confined to the vehicle domain, which may limit its generalizability to other areas such as electronics or healthcare. 
However, since the benchmark construction pipeline, which includes data collection, tuple annotation, and insight generation, is designed to be domain-agnostic, it can be easily extended to other fields with only minor adjustments to data sourcing and annotation guidelines.

Second, the current benchmark does not take into account user-specific information (user profiles). 
In real-world applications, factors such as user expertise, preferences, usage context, and prior sentiment trends play a critical role in shaping actionable insights. 
Integrating user metadata or interaction history would enable a natural extension of the framework toward user-aware opinion mining. 
While this direction is beyond the current scope, enriching \proposed with such annotations and modeling could open up new avenues for personalized opinion mining, allowing LLMs to produce more tailored and context-sensitive outputs.

Third, although we adopted a human–machine collaborative annotation pipeline~\cite{sharif2024explicit, seo2025mt} to construct high-quality labels, opinion extraction and summarization inherently involve subjective judgment. 
To mitigate this, we established detailed annotation guidelines and a multi-stage validation process; nevertheless, some degree of annotation variance is unavoidable. 
Future work may explore more systematic approaches to subjectivity, such as crowdsourced consensus annotation, uncertainty-aware learning frameworks, or prompt ensemble methods.

%% file: 090ethical_statement.tex

This study strictly adhered to ethical guidelines throughout the entire process of data collection and usage. 
As the benchmark is limited to the automotive domain, concerns may arise regarding representativeness and potential amplification of biases related to gender, socioeconomic status, or geography. 
To address this issue, we collected data from global platforms such as Reddit, YouTube and well-known review blogs which are widely used by diverse communities across different regions and demographics. 
We believe this design helps capture a broad spectrum of opinions present in real-world online discourse, though we acknowledge that residual biases may still persist. 

Regarding data collection procedures, crawling was conducted solely for non-commercial research purposes and performed at a controlled rate to prevent server overload or potential DDoS attacks.
When collecting user reviews, personal information such as reviewer IDs, names, and locations was intentionally excluded, focusing only on textual content to ensure user privacy. 
However, we cannot completely rule out the possibility that review texts may contain sensitive personal details, hate speech, or inappropriate content. 

Finally, all data samples were collected and annotated in compliance with the terms and conditions of their respective sources. 
By releasing our dataset, we aim to contribute to the academic advancement of generative opinion mining research.

%% file: 100acknowledgements.tex
This work was supported by the IITP grants funded by the Korea government (MSIT) (No. RS-2020-
II201361; RS-2024-00457882, AI Research Hub Project), and the NRF grant funded by the Korea
government (MSIT) (No. RS-2025-00560295).

%% file: 081appendix.tex
\section{Benchmark Construction Details}
\label{sec:oomb_details}
\subsection{Data Source}
\label{subsec:data_source_details}
To construct a diverse and representative dataset for online opinion mining, we collected user-generated content from four distinct web content types: \textit{Blog}, \textit{Reddit}, \textit{Review Site}, and \textit{YouTube}.


\paragraph{Blog}  
We collected data from well-established automotive blogs, including The Drive\footnote{\url{https://www.thedrive.com/category/car-reviews}} , Autoblog\footnote{\url{https://www.autoblog.com/reviews/}}, and CarExpert.\footnote{\url{https://www.carexpert.com.au/car-reviews}} 
Blog content is primarily written by experts and car owners, often providing detailed and comprehensive insights on a single entity. 
Compared to other content types, blog posts tend to be longer and more structured, covering multiple aspects of a vehicle in depth.

\paragraph{Reddit} 
We collected data from the r/cars subreddit\footnote{\url{https://www.reddit.com/r/cars/}} , a community of car enthusiasts. Users freely share their opinions about various vehicles through a multi-threaded structure, where multiple participants engage in open discussions. 
This interactive nature generates diverse automotive perspectives through community discussions, making it a valuable source for opinion mining.

\paragraph{Review Site} 
We collected data from Edmunds\footnote{\url{https://www.edmunds.com/car-reviews/}} , an automotive review platform where users provide star ratings along with detailed reviews for specific vehicles. 
Review sites explicitly encourage opinion sharing, leading to more direct and detailed user feedback. 
These structured reviews combine ratings with detailed feedback, making them rich in straightforward user opinions.

\paragraph{YouTube}  
We gathered comments from automotive YouTube channels\footnote{\url{https://www.youtube.com/@AutoTraderTV}} \footnote{\url{https://www.youtube.com/channel/UCsqjHFMB_JYTaEnf_vmTNqg}}  listed by Feedspot\footnote{\url{https://videos.feedspot.com/car_youtube_channels/}}, focusing on channels with large subscriber bases providing car reviews and analysis. 
When YouTubers share their vehicle reviews, viewer opinions and reactions appear in the comments. 
YouTube uses a single-threaded structure where viewers can leave comments and engage in discussions through replies. 
This structure allows for community participation through viewer responses to both the video content and other comments, creating an interactive space for opinion sharing.






\subsection{Data Annotation Details}
\label{subsec:appendix_annotation_refinement}
\paragraph{Human-in-the-loop annotation}
To ensure high-quality, consistent annotations, we adopted a human-in-the-loop process in which \gptmini\footnote{\texttt{gpt-4o-mini-2024-07-18}} serves as the initial annotator and human annotators\footnote{We recruit undergraduates and graduates who are proficient in English and knowledgeable in the automotive domain.} subsequently verify and refine its outputs.  
Prompts used to solicit these initial annotations are provided in Table~\ref{tab:tuple_annotation_prompt} for tuples and Table~\ref{tab:summary_annotation_prompt} for insights.  
All annotators underwent one full week of training on our detailed guidelines (Table~\ref{tab:tuple_guideline} for tuples; Table~\ref{tab:summary_guideline} for insights) before beginning any annotation work. 
This entire annotation–refinement process was applied to every sample in the dataset, ensuring higher overall quality and consistency in the resulting annotations.

\paragraph{Inter-Annotator Agreement Details}
We provide additional details of the inter-annotator agreement evaluation reported in Table~\ref{tab:iaa_results}.
\begin{itemize}[leftmargin=*,topsep=4pt,itemsep=4pt,parsep=0pt]
    \item \textbf{Span-level tuple agreement}: Each annotator independently extracted (entity, feature, opinion) tuples from every sentence in the selected contents. 
    Inter-annotator consistency was measured using Fleiss’ Kappa, yielding a score of 0.6821, which generally indicates substantial agreement among multiple annotators.
    This result indicates strong inter-annotator reliability in capturing core opinion structures at the span level.
    
    \item \textbf{Preferred insight agreement}: For each content, annotators selected their top-1 to top-3 preferred opinion-centric insights from five LLM-generated candidate insights.
    We computed Fleiss’ Kappa over these preference rankings and obtained a score of 0.7695, indicating substantial consistency among annotators.

    \item \textbf{Insight semantic consistency}: After guideline-based verification and rewriting, we computed the pairwise semantic similarity between the finalized opinion-centric insights produced by different annotators. 
    We used BERTScore to measure semantic alignment, and report the average score across all annotator pairs, which reached a very high value of 0.9297. 
    This indicates that, despite differences in wording, annotators consistently expressed the same underlying meaning.
\end{itemize}

\paragraph{Annotation Tools}
To enhance clarity and efficiency in the annotation process, we developed dedicated UIs for each stage of verification and refinement.
Tuple verification and refinement were performed via our annotation UI (see Figure \ref{fig:tuple_ui_1} and \ref{fig:tuple_ui_2}), which displays the original content, each tuple’s components with existence flags, and highlighted evidence sentences. 
Insight verification and refinement were conducted using a separate annotation UI (see Figure \ref{fig:summary_ui}), which presents the content text, the associated gold tuple, and the working draft of the insight side by side for comparison and iterative improvement.

\subsection{Dataset Statistics}
\label{subsec:dataset_statistics}
We provide detailed statistics on key attributes—namely, the number of samples, average token count, number of users, and number of tuples (i.e., opinions)—for each of the four content types collected: blogs, Reddit, review sites, and YouTube.
As shown in Table~\ref{tab:oomb_stats}, we categorize the values of each attribute into predefined ranges to illustrate the distribution of samples across different levels.
Token lengths were measured using the NLTK word\_tokenize library\footnote{\url{https://www.nltk.org/api/nltk.tokenize.word_tokenize.html}}.

\subsection{Dataset Analysis}
\label{subsec:data_topic_analysis}
\begin{figure*}[t]
\centering
\includegraphics[width=\textwidth]{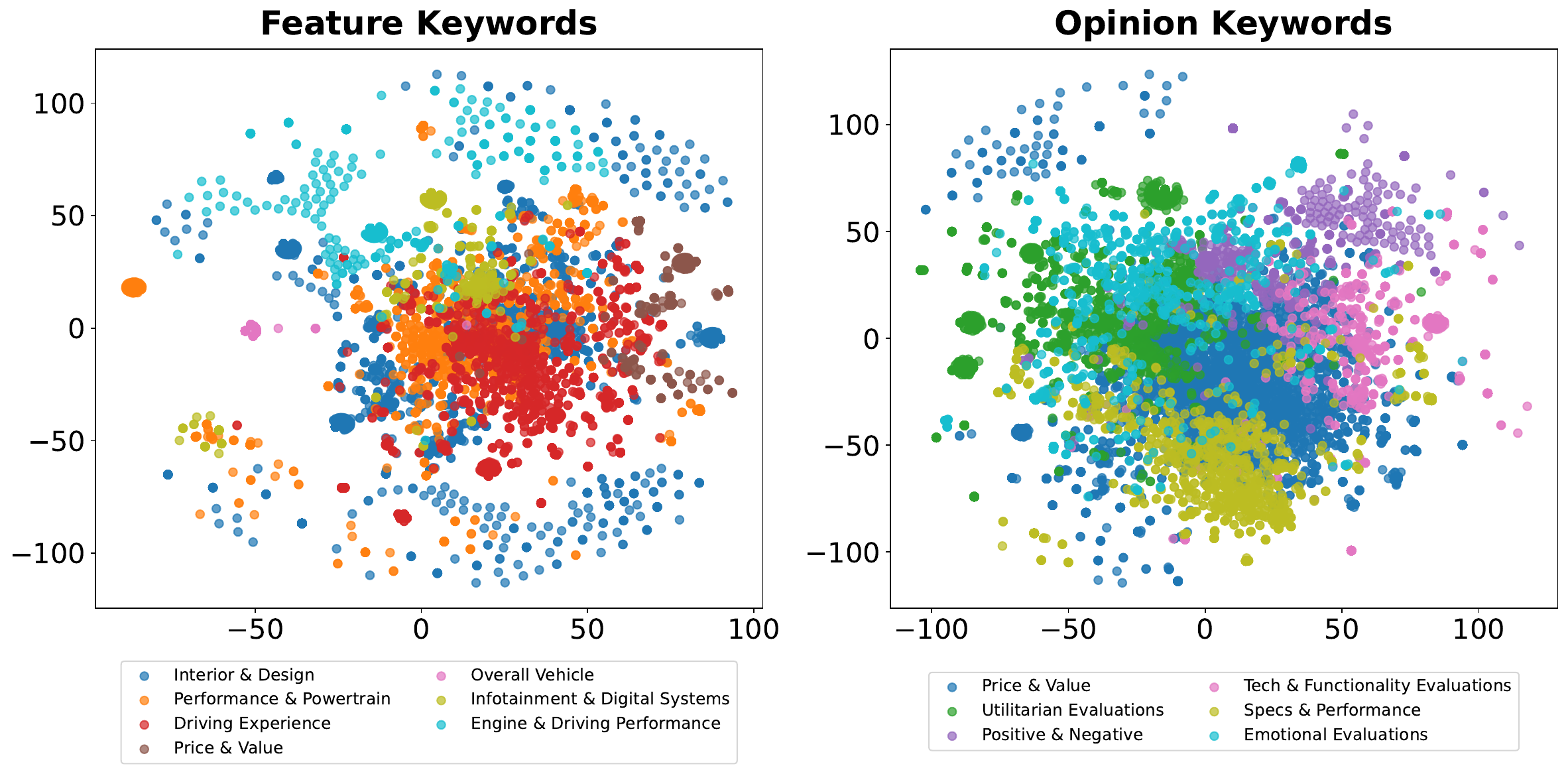}
\caption{Visualization of feature keywords \textbf{(Left)} and opinion keywords \textbf{(Right)} extracted via K-means clustering using t-SNE. 
Each point represents a keyword, and colors indicate different topic clusters.}
\label{fig:topic_analysis}
\end{figure*}

Additionally, Figure~\ref{fig:topic_analysis} provides the distribution of feature-opinion topics in our dataset.
\paragraph{Feature keywords}
The t-SNE visualization shows that feature keywords form well-separated clusters according to major product aspects. 
Categories like \textit{Interior \& Design}, \textit{Driving Experience}, and \textit{Performance \& Powertrain} appear frequently and show high cohesion. 
\textit{Engine \& Driving Performance} and \textit{Infotainment \& Digital Systems} are located near the core clusters, reflecting semantic proximity, while Overall Vehicle appears more scattered, indicating higher lexical variability. 
Outlier points on the periphery suggest rare or ambiguous feature expressions that may require finer handling.

\paragraph{Opinion keywords}
Opinion keywords also exhibit meaningful clustering patterns. \textit{Price \& Value} and \textit{Specs \& Performance} form tight clusters, while \textit{Utilitarian and Emotional Evaluations} overlap in the center, suggesting a blend of practical and emotional judgments. 
\textit{Tech \& Functionality} Evaluations appear in a distinct region, separate from general \textit{Positive \& Negative} sentiment expressions, highlighting their specialized nature. 
Points between clusters reflect nuanced or polysemous opinions, suggesting the need for flexible sentiment understanding models.



\section{Experimental Details}
\label{sec:experimental_details}
\subsection{Evaluation Models}


\paragraph{Proprietary LLMs}
We used the most up-to-date versions of OpenAI APIs\footnote{\url{https://openai.com/index/openai-api/}} and Anthropic AI\footnote{\url{https://www.anthropic.com/}}.  
Specifically, we used the following models:

\begin{itemize}[leftmargin=*,topsep=4pt,itemsep=4pt,parsep=0pt]
    \item \textbf{\gptmini}: {\fontfamily{lmtt}\selectfont gpt-4o-mini-2024-07-18} 
    \item \textbf{\gptfouro}: {\fontfamily{lmtt}\selectfont gpt-4o-2024-08-06} 
    \item \textbf{\claudehaiku}: {\fontfamily{lmtt}\selectfont claude-3-5-haiku-20241022}
    \item \textbf{\claudesonnet}: {\fontfamily{lmtt}\selectfont claude-3-5-sonnet-20241022}
\end{itemize}

\paragraph{Open-source LLMs}
We used Hugging Face model cards and ran them on two NVIDIA A100 GPUs. 
Specifically, we used the following models:

\begin{itemize}[leftmargin=*,topsep=4pt,itemsep=4pt,parsep=0pt]
    \item \textbf{\llamaeight}: {\fontfamily{lmtt}\selectfont meta-llama/meta-llama-3-8b-instruct}
    \item \textbf{\llamaseventy}: {\fontfamily{lmtt}\selectfont meta-llama/meta-llama-3-70b-instruct}
    \item \textbf{\gemmanine}: {\fontfamily{lmtt}\selectfont google/gemma-2-9b-it}
    \item \textbf{\gemmatwentyseven}: {\fontfamily{lmtt}\selectfont google/gemma-2-27b-it}
    \item \textbf{\qwen}: {\fontfamily{lmtt}\selectfont Qwen/Qwen2.5-7B-Instruct}
    \item \textbf{\deepseek}: {\fontfamily{lmtt}\selectfont deepseek-ai/deepseek-llm-7b-chat}
\end{itemize}

\subsection{\subone}
\label{subsec:appendix_foe}
\paragraph{RM metric threshold selection}
\label{para:rm_abalation_study}
We set the threshold of the Relaxed Match (RM) metric to 0.7, as it empirically provides the optimal balance—capturing meaningful semantic similarities without being overly permissive.
Prior information extraction (IE) research has highlighted that exact span matching may underestimate model performance due to its overly strict nature. 
To alleviate this issue, previous studies have proposed overlap-based evaluations with thresholds set at 0.5 \cite{han2023information} or 0.75 \cite{sharif2024explicit}.

As shown in Table \ref{tab:l_rm_abalation} and \ref{tab:s_rm_abalation}, lower thresholds tend to excessively acknowledge partial overlaps, inflating recall to an unrealistic degree. 
Conversely, higher thresholds often miss semantically valid matches due to minor textual variations, causing the RM metric performance to converge toward Exact Match (EM) scores and consequently lose its intended flexibility.
Thus, our experiments confirm that a threshold of 0.7 achieves optimal RM performance, which we subsequently adopt for our main experiments.

\paragraph{Human Alignment in Tuple Matching}
\label{para:correlation}
This experiment aims to identify the most appropriate matching metric for reliably evaluating LLMs’ tuple extraction performance. 
To this end, we investigate which of the three evaluation metrics used in the FOE task—Exact Match (EM), Relaxed Match (RM), and Contextual Match (CM)—best aligns with human judgment.
First, we randomly select 100 tuples predicted by GPT-4o given the input content texts. 
Then, five human annotators evaluated the validity of each predicted-gold tuple pair using binary judgments: 1 if they considered the pair to be a match, and 0 otherwise. Based on these judgments, we computed the correlation coefficients between human agreement and each metric.

Table~\ref{tab:correlation_results} reports the Pearson \(r\), Spearman \(\rho\), and Kendall \(\tau\) correlation coefficients, averaged across the five annotators over the 100 samples. 
Across all correlation metrics, CM achieved the highest alignment with human judgment, indicating that Contextual Match best reflects how humans assess tuple matching.
These findings suggest that CM serves as the most reliable and appropriate metric for evaluating the performance of LLMs in tuple extraction tasks.

\subsection{\subtwo}
\label{subsec:appendix_oig}
\paragraph{LLM-as-a-judge evaluation criteria}
\label{para:LLM_judge_eval_criteria}
To evaluate insight quality across diverse criteria, we use \gptfouro\ and randomly sample 100 pieces of content from each of the four content types.
The scoring scale for each evaluation follows the previous NLG evaluation framework, G-EVAL \cite{liu2023gevalnlgevaluationusing}, and is measured on a scale from 1 to 5.
We adopt the following six criteria:

\begin{itemize}[leftmargin=*,topsep=4pt,itemsep=4pt,parsep=0pt]
    \item \textbf{Faithfulness}: Evaluate whether the insight faithfully reflects the original review without distortion and check for any hallucinations.
    \item \textbf{Coverage}: Evaluate whether the insight effectively captures and represents the key opinions expressed in the review.
    \item \textbf{Specificity}: Evaluate whether the insight presents meaningful and relevant details rather than being vague or overly generic.
    \item \textbf{Insightfulness} : Evaluate whether the insight offers meaningful observations or interpretations that enhance understanding or decision-making for the reader.
    \item \textbf{Intent}: Evaluate whether the insight accurately preserves the author's original tone, intent, and nuances without altering the emotional or stylistic essence of the review.
    \item \textbf{Fluency}: Evaluate whether the insight is naturally written, grammatically correct, and easy to read.
\end{itemize}

\paragraph{Case study: Annotated opinion-centric insight}
\label{para:oig_case_study_gold_summary}
Table~\ref{tab:oig_gold_summary_case_study2} presents representative examples of gold opinion-centric insights for each content type.

\paragraph{Case study: LLMs’ insight generation capability}
\label{para:oig_case_study_g_eval}
We illustrate our LLM-as-a-judge evaluation protocol with two case studies: one on a review site (Tables~\ref{tab:llm_judge_eval_case_study_11} and~\ref{tab:llm_judge_eval_case_study_12}), and one on a Reddit (Tables~\ref{tab:llm_judge_eval_case_study_21} and~\ref{tab:llm_judge_eval_case_study_22}). 
In each set, the first table shows the input content, the model-generated insight (\gptfouro), the gold opinion-centric insight, and the automatic metric A3CU score, while the second table provides a reference-free, six-dimension LLM-as-a-judge assessment complete with per-dimension scores and detailed reasoning.  

Although the A3CU metric assigns the review site example a high score (60.91) and the Reddit example a low score (10.12), our reference-free LLM-as-a-judge evaluation reveals that the model’s performance on the Reddit content is in fact stronger across several human-aligned dimensions—particularly \textit{Coverage} and \textit{Specificity}.
This divergence highlights the limitations of purely reference-based, automatic metrics in capturing the nuanced, insight-oriented qualities of opinion insights. 
We therefore conclude that for OIG evaluation, combining automatic reference-based metrics with a reference-free, human-aligned judging protocol yields a more comprehensive and reliable assessment of LLMs’ true insight‐generation capabilities.  

\paragraph{Human evaluation}
\label{para:amazon_hman_eval}
We assess the quality of the generated insight through a human evaluation conducted on Amazon Mechanical Turk (AMT).
Specifically, we randomly sample 200 examples from our benchmark and ask three human judges per example to evaluate insights generated by GPT-4o under two settings: 1) using only the input content, and 2) using both the input content and gold tuples.
Each judge rates the quality of the insights on a 1 to 5 scale across six criteria.
The AMT interface used for human evaluation is presented in Figure \ref{fig:humaneval_main}, \ref{fig:humaneval_faithfulness}, \ref{fig:humaneval_coverage}, \ref{fig:humaneval_specificity}, \ref{fig:humaneval_insightfulness}, \ref{fig:humaneval_intent} and \ref{fig:humaneval_fluency}.

\subsection{Performance of LLMs by content type}
\label{sec:additaionl_results}
We report the \subone performance of various LLMs for each content type in Table \ref{tab:subone_blog}, \ref{tab:subone_reddit}, \ref{tab:subone_review_site} and \ref{tab:subone_youtube} , and the \subtwo performance in Table \ref{tab:oig_blog_reddit}, \ref{tab:oig_review_youtube}.

\subsection{Prompts}
We present prompts used in our experiments:
\begin{itemize}
    \item \textbf{Data Annotation}: The prompt designed for \subone is shown in Table \ref{tab:tuple_annotation_prompt} and the prompt for \subtwo is shown in Table~\ref{tab:summary_annotation_prompt} 
    \item \textbf{\subone (FOE)}: The prompt designed for \subone is shown in Table \ref{tab:subone_prompt}.
    \item \textbf{\subtwo (OIG)}: The prompt designed for \subtwo is shown in Table \ref{tab:subtwo_prompt}.
    \item \textbf{Contextual Match (CM)}: The prompt used for performing Contextual Match is provided in Table \ref{tab:gm_prompt}.
    \item \textbf{LLM-as-a-judge Evaluation}: The prompt for evaluating the LLM-Judge is presented in Table \ref{tab:llm_judge_faithfulness}, \ref{tab:llm_judge_coverage}, \ref{tab:llm_judge_specificity}, \ref{tab:llm_judge_insightfulness}, \ref{tab:llm_judge_intent} and \ref{tab:llm_judge_fluency}.
\end{itemize}

\input{TAB/084data_statistic}

\input{TAB/085rm_ablation_study}
\input{TAB/081extractive}

\input{TAB/082abstractive}
\input{TAB/083prompt}

\input{TAB/087annotatoin_guidlines}
\input{TAB/086summary_case_study}

\input{TAB/086summary_case_study_g_eval}

\clearpage
\newpage

\input{TAB/089annotation_ui}

\input{TAB/088human_eval_amt}

%% file: TAB/084data_statistic.tex
\begin{table*}[!t]
  \centering
  \footnotesize  
  \setlength{\tabcolsep}{6pt}  
  \begin{tabular}{
    l
    l
    r
    r
    r
    r
    r
    r}
    \toprule
    \multirow{2}{*}[-0.5ex]{\textbf{Attribute}}
      & \multirow{2}{*}[-0.5ex]{\textbf{Range}}
      & \multicolumn{4}{c}{\textbf{Content Type}}
      & \multirow{2}{*}[-0.5ex]{\textbf{Total}}
      & \multirow{2}{*}[-0.5ex]{\shortstack{\textbf{Ratio}\textbf{(\%)}}} \\
    \cmidrule(lr){3-6}
      & & \textbf{Blog} & \textbf{Reddit} & \textbf{Review Site} & \textbf{YouTube}
      & & \\
    \midrule
    
    \multirow{6}{*}{\textbf{Tuples}}
      & $\leq 10$   & \phantom{0}8 & \phantom{0}79 & 14 & 125 & 226 & 37.7 \\
      & $\leq 15$   & 14 & 49 & 33 & \phantom{0}32 & 128 & 21.3 \\
      & $\leq 20$   & 19 & 21 & 33 & \phantom{0}21 & \phantom{0}94 & 15.7 \\
      & $\leq 25$   & 20 & 15 & 13 & \phantom{00}7 & \phantom{0}55 & \phantom{0}9.2 \\
      & $\leq 30$   & 20 & \phantom{0}8 & \phantom{0}8 & \phantom{00}8 & \phantom{0}44 & \phantom{0}7.3 \\
      & $\leq 35$   & 34 & \phantom{0}5 & \phantom{0}5 & \phantom{00}9 & \phantom{0}53 & \phantom{0}8.8 \\
    \midrule
    \multirow{6}{*}{\textbf{Entities}}
      & 1         & 58 & 53 & 89 & 128 & 328 & 54.7 \\
      & 2         & 44 & 34 & 12 & \phantom{0}39 & 129 & 21.5 \\
      & 3         & \phantom{0}9 & 24 & \phantom{0}3 & \phantom{0}17 & \phantom{0}53 & \phantom{0}8.8 \\
      & 4         & \phantom{0}2 & 24 & \phantom{0}0 & \phantom{00}7 & \phantom{0}33 & \phantom{0}5.5 \\
      & 5         & \phantom{0}1 & 10 & \phantom{0}1 & \phantom{00}3 & \phantom{0}15 & \phantom{0}2.5 \\
      & $\geq 6$  & \phantom{0}1 & 32 & \phantom{0}1 & \phantom{00}8 & \phantom{0}42 & \phantom{0}7.0 \\
    \midrule
    \multirow{5}{*}{\textbf{Tokens}}
      & $\leq 200$   & \phantom{00}0 & 100 & \phantom{0}0 & 81 & 181 & 30.2 \\
      & $\leq 400$   & \phantom{00}3 & \phantom{0}58 & 85 & 68 & 214 & 35.7 \\
      & $\leq 1000$  & \phantom{00}9 & \phantom{0}14 & 21 & 44 & \phantom{0}88 & 14.7 \\
      & $\leq 2000$  & 39 & \phantom{00}3 & \phantom{0}0 & \phantom{0}8 & \phantom{0}50 & \phantom{0}8.3 \\
      & $\leq 3000$  & 64 & \phantom{00}2 & \phantom{0}0 & \phantom{0}1 & \phantom{0}67 & 11.2 \\
    \midrule
    \multirow{5}{*}{\textbf{Users}}
      & 1         & 58 & 47 & 89 & 128 & 322 & 53.7 \\
      & 2         & 44 & 35 & 12 & \phantom{0}39 & 130 & 21.7 \\
      & 3         & \phantom{0}9 & 24 & \phantom{0}3 & \phantom{0}17 & \phantom{0}53 & \phantom{0}8.8 \\
      & 4         & \phantom{0}2 & 26 & \phantom{0}0 & \phantom{00}7 & \phantom{0}35 & \phantom{0}5.8 \\
      & $\geq 5$  & \phantom{0}2 & 45 & \phantom{0}2 & \phantom{0}11 & \phantom{0}60 & 10.0 \\
    \midrule
    \multicolumn{2}{c}{\textbf{Total}}
      & 115 & 177 & 106 & 202 & 600 & 100.0 \\
    \bottomrule
  \end{tabular}
  \caption{Statistics of the \proposed~dataset across different attributes.}
  \label{tab:oomb_stats}
\end{table*}

%% file: TAB/085rm_ablation_study.tex
\begin{table*}[h]
\centering
\small
\resizebox{0.99\linewidth}{!}{
\begin{tabular}{lcccccccccccc}
\toprule
\multirow{2.5}{*}{\textbf{Models}}
& \multicolumn{3}{c}{\textbf{EM}} 
& \multicolumn{3}{c}{\textbf{L-RM\,$\ge0.7$}}
& \multicolumn{3}{c}{\textbf{L-RM\,$\ge0.8$}}
& \multicolumn{3}{c}{\textbf{L-RM\,$\ge0.9$}} \\
\cmidrule(lr){2-4}\cmidrule(lr){5-7}\cmidrule(lr){8-10}\cmidrule(lr){11-13}
& \textbf{Pre} & \textbf{Rec} & \textbf{F1} 
& \textbf{Pre} & \textbf{Rec} & \textbf{F1} 
& \textbf{Pre} & \textbf{Rec} & \textbf{F1} 
& \textbf{Pre} & \textbf{Rec} & \textbf{F1} \\ \midrule
\multicolumn{13}{c}{\textbf{\textit{Proprietary LLMs}}} \\ \midrule
\gptmini          & 3.91 & 1.99 & 2.62 & 11.29 & 5.70 & 7.52 & 7.72 & 3.68 & 4.98 & 4.80 & 2.29 & 3.10 \\  
\gptfouro         & \underline{7.27} & \underline{5.18} & \underline{6.02} & \underline{15.86} & \underline{11.39} & \underline{13.20} & \underline{11.90} & \underline{8.51} & \underline{9.92} & \underline{8.43} & \underline{6.03} & \underline{7.03} \\  
\claudehaiku      & 6.13 & 3.02 & 4.01 & 15.02 & 7.51 & 9.94 & 10.61 & 4.97 & 6.77 & 7.17 & 3.36 & 4.57 \\  
\claudesonnet     & \textbf{11.12} & \textbf{6.32} & \textbf{7.97} & \textbf{22.97} & \textbf{13.02} & \textbf{16.46} & \textbf{17.81} & \textbf{9.61} & \textbf{12.48} & \textbf{12.57} & \textbf{6.78} & \textbf{8.81} \\ \midrule
\multicolumn{13}{c}{\textbf{\textit{Open-source LLMs}}} \\ \midrule
\llamaeight       & \textbf{8.49} & \textbf{6.28} & \textbf{7.17} & \textbf{16.75} & \textbf{12.18} & \textbf{14.02} & \textbf{13.13} & \textbf{9.17} & \textbf{10.80} & \textbf{9.85} & \textbf{6.88} & \textbf{8.10} \\  
\llamaseventy     & \underline{7.26} & 5.57 & \underline{6.27} & \underline{15.21} & 11.66 & \underline{13.13} & \underline{12.07} & 9.07 & \underline{10.36} & \underline{8.67}& \underline{6.52} & \underline{7.44} \\  
\gemmanine        & 6.37 & 4.51 & 5.25 & 14.17 & 10.17 & 11.78 & 9.98 & 7.27 & 8.41 & 7.11 & 5.18 & 5.99 \\  
\gemmatwentyseven & 7.05 & \underline{5.61} & 6.20 & 14.33 & \underline{11.77} & 12.82 & 10.80 & \underline{9.08} & 9.87 & 7.55 & 6.35 & 6.90 \\    
\qwen             & 6.55 & 4.18 & 5.05 & 12.97 & 8.32 & 10.02 & 9.99 & 6.14 & 7.60 & 7.35 & 4.51 & 5.59 \\  
\deepseek         & 3.00 & 1.63 & 2.07 & 5.86 & 3.13 & 4.02 & 4.88 & 2.31 & 3.14 & 3.78 & 1.79 & 2.43 \\   
\bottomrule
\end{tabular}
}
\caption{Ablation study on the FOE task using three L-RM thresholds ($\ge$0.7, $\ge$0.8, $\ge$0.9).}
\label{tab:l_rm_abalation}
\end{table*}

\begin{table*}[h]
\centering
\small
\resizebox{0.99\linewidth}{!}{
\begin{tabular}{lcccccccccccc}
\toprule
\multirow{2.5}{*}{\textbf{Models}}
& \multicolumn{3}{c}{\textbf{EM}} 
& \multicolumn{3}{c}{\textbf{S-RM\,$\ge0.7$}}
& \multicolumn{3}{c}{\textbf{S-RM\,$\ge0.8$}}
& \multicolumn{3}{c}{\textbf{S-RM\,$\ge0.9$}} \\
\cmidrule(lr){2-4}\cmidrule(lr){5-7}\cmidrule(lr){8-10}\cmidrule(lr){11-13}
& \textbf{Pre} & \textbf{Rec} & \textbf{F1} 
& \textbf{Pre} & \textbf{Rec} & \textbf{F1} 
& \textbf{Pre} & \textbf{Rec} & \textbf{F1} 
& \textbf{Pre} & \textbf{Rec} & \textbf{F1} \\ \midrule
\multicolumn{13}{c}{\textbf{\textit{Proprietary LLMs}}} \\ \midrule
\gptmini          & 3.91 & 1.99 & 2.62 & 15.50 & 7.76 & 10.27 & 9.86 & 4.70 & 6.37 & 5.48 & 2.61 & 3.54 \\  
\gptfouro         & \underline{7.27} & \underline{5.18} & \underline{6.02} & \underline{21.23} & \underline{15.31} & \underline{17.71} & \underline{14.76} & \underline{10.55} & \underline{12.31} & \underline{9.39} & \underline{6.72} & \underline{7.83} \\  
\claudehaiku      & 6.13 & 3.02 & 4.01 & 20.60 & 10.36 & 13.68 & 13.41 & 6.28 & 8.56 & 8.55 & 4.00 & 5.45 \\  
\claudesonnet     & \textbf{11.12} & \textbf{6.32} & \textbf{7.97} & \textbf{29.30} & \textbf{16.52} & \textbf{20.93} & \textbf{21.08} & \textbf{11.37} & \textbf{14.78} & \textbf{14.08} & \textbf{7.59} & \textbf{9.87} \\ \midrule
\multicolumn{13}{c}{\textbf{\textit{Open-source LLMs}}} \\ \midrule
\llamaeight       & \textbf{8.49} & \textbf{6.28} & \textbf{7.17} & \textbf{21.33} & \underline{15.43} & \textbf{17.80} & \textbf{15.52} & \underline{10.84} & \textbf{12.77} & \textbf{10.59} & \textbf{7.39} & \textbf{8.71} \\  
\llamaseventy     & \underline{7.26} & 5.57 & \underline{6.27} & \underline{19.94} & 15.18 & 17.15 & \underline{13.80} & 10.37 & 11.85 & \underline{9.44} & \underline{7.09} & \underline{8.10} \\  
\gemmanine        & 6.37 & 4.51 & 5.25 & 17.73 & 12.59 & 14.64 & 11.73 & 8.55 & 9.89 & 7.73 & 5.64 & 6.52 \\  
\gemmatwentyseven & 7.05 & \underline{5.61} & 6.20 & 19.29 & \textbf{15.83} & \underline{17.25} & 13.09 & \textbf{11.01} & \underline{11.96} & 8.18 & 6.88 & 7.48 \\    
\qwen             & 6.55 & 4.18 & 5.05 & 18.22 & 11.83 & 14.18 & 12.36 & 7.59 & 9.41 & 8.38 & 5.15 & 6.38 \\  
\deepseek         & 3.00 & 1.63 & 2.07 & 8.20 & 4.34 & 5.61 & 5.68 & 2.69 & 3.65 & 3.94 & 1.87 & 2.53 \\   
\bottomrule
\end{tabular}
}
\caption{Ablation study on the FOE task using three S-RM thresholds ($\ge$0.7, $\ge$0.8, $\ge$0.9).}
\label{tab:s_rm_abalation}
\end{table*}

%% file: TAB/081extractive.tex
\begin{table*}[t]
\centering
\small
\resizebox{0.99\linewidth}{!}{
\begin{tabular}{lcccccccccccc}
\toprule
\multirow{2.5}{*}{\textbf{Models}}
& \multicolumn{3}{c}{\textbf{EM}} 
& \multicolumn{3}{c}{\textbf{L-RM}} 
& \multicolumn{3}{c}{\textbf{S-RM}} 
& \multicolumn{3}{c}{\textbf{CM}} \\
\cmidrule(lr){2-4}\cmidrule(lr){5-7}\cmidrule(lr){8-10}\cmidrule(lr){11-13}
& \textbf{Pre} & \textbf{Rec} & \textbf{F1} 
& \textbf{Pre} & \textbf{Rec} & \textbf{F1} 
& \textbf{Pre} & \textbf{Rec} & \textbf{F1} 
& \textbf{Pre} & \textbf{Rec} & \textbf{F1} \\ \midrule
\multicolumn{13}{c}{\textbf{\textit{Proprietary LLMs}}} \\ \midrule
\gptmini & 2.40 & 1.08 & 1.49 & 10.17 & 4.58 & 6.31 & 13.44 & 6.05 & 8.34 & \textbf{49.14} & 22.55 & 29.49 \\  
\gptfouro & \underline{4.67} & \textbf{3.53} & \textbf{4.02} & \underline{14.27} & \textbf{10.78} & \underline{12.29} & \underline{17.91} & \textbf{13.53} & \underline{15.41} & 38.21 & \textbf{28.34} & \textbf{30.63} \\ 
\claudehaiku & 3.31 & 1.44 & 2.00 & 10.15 & 4.41 & 6.15 & 14.96 & 6.50 & 9.07 & \underline{46.95} & 21.98 & 28.75 \\  
\claudesonnet & \textbf{5.68} & \underline{2.91} & \underline{3.85} & \textbf{18.75} & \underline{9.61} & \textbf{12.71} & \textbf{25.51} & \underline{13.07} & \textbf{17.29} & 45.30 & \underline{23.97} & \underline{30.08} \\ \midrule
\multicolumn{13}{c}{\textbf{\textit{Open-source LLMs}}} \\ \midrule
\llamaeight & \underline{4.36} & 2.78 & \underline{3.39} & \underline{11.74} & 7.48 & 9.14 & \underline{14.82} & 9.44 & \underline{11.54} & \textbf{39.53} & 25.06 & \textbf{29.20} \\ 
\llamaseventy & \textbf{5.86} & \textbf{4.41} & \textbf{5.04} & \textbf{12.34} & \textbf{9.28} & \textbf{10.59} & \textbf{15.86} & \textbf{11.93} & \textbf{13.61} & \underline{35.42} & \underline{27.00} & \underline{28.96} \\
\gemmanine & 3.38 & \underline{2.80} & 3.06 & 10.01 & \underline{8.36} & \underline{9.15} & 11.98 & 9.92 & 10.86 & 33.45 & 26.90 & 27.64 \\
\gemmatwentyseven & 2.54 & 2.54 & 2.54 & 8.19 & 8.22 & 8.20 & 10.95 & \underline{10.98} & 10.96 & 30.76 & \textbf{27.42} & 26.27 \\  
\qwen & 3.81 & 1.96 & 2.59 & 8.83 & 4.54 & 6.00 & 12.95 & 6.67 & 8.80 & 28.15 & 18.26 & 20.24 \\ 
\deepseek & 1.57 & 0.49 & 0.75 & 2.72 & 0.85 & 1.29 & 3.24 & 1.01 & 1.54 & 34.89 & 12.07 & 16.25 \\ 
\bottomrule
\end{tabular}
}
\caption{Performance comparison of different models for FOE task  across various evaluation metrics on Blog type.}
\label{tab:subone_blog}
\end{table*}

\begin{table*}[t]
\centering
\small
\resizebox{0.99\linewidth}{!}{
\begin{tabular}{lcccccccccccc}
\toprule
\multirow{2.5}{*}{\textbf{Models}}
& \multicolumn{3}{c}{\textbf{EM}} 
& \multicolumn{3}{c}{\textbf{L-RM}} 
& \multicolumn{3}{c}{\textbf{S-RM}} 
& \multicolumn{3}{c}{\textbf{CM}} \\
\cmidrule(lr){2-4}\cmidrule(lr){5-7}\cmidrule(lr){8-10}\cmidrule(lr){11-13}
& \textbf{Pre} & \textbf{Rec} & \textbf{F1} 
& \textbf{Pre} & \textbf{Rec} & \textbf{F1} 
& \textbf{Pre} & \textbf{Rec} & \textbf{F1} 
& \textbf{Pre} & \textbf{Rec} & \textbf{F1} \\ \midrule
\multicolumn{13}{c}{\textbf{\textit{Proprietary LLMs}}} \\ \midrule
\gptmini & 3.65 & 1.43 & 2.06 & 9.23 & 3.62 & 5.20 & 14.30 & 5.61 & 8.05 & \underline{68.73} & 33.07 & 41.11 \\
\gptfouro & \underline{9.74} & \textbf{5.65} & \underline{7.15} & \underline{18.18} & \textbf{10.54} & \textbf{13.34} & \underline{23.87} & \textbf{13.84} & \textbf{17.52} & 63.59 & 41.83 & \textbf{47.70} \\
\claudehaiku & 6.25 & 2.43 & 3.49 & 14.34 & 5.57 & 8.02 & 18.34 & 7.12 & 10.25 & \textbf{68.80} & 34.67 & 42.43 \\
\claudesonnet & \textbf{13.14} & \underline{5.41} & \textbf{7.66} & \textbf{22.32} & \underline{9.18} & \underline{13.01} & \textbf{28.60} & \underline{11.77} & \underline{16.68} & 68.68 & 34.84 & \underline{42.90} \\ \midrule
\multicolumn{13}{c}{\textbf{\textit{Open-source LLMs}}} \\ \midrule
\llamaeight & \underline{9.28} & \underline{5.61} & \underline{6.99} & \textbf{17.31} & \textbf{10.46} & \textbf{13.04} & \textbf{22.45} & \textbf{13.56} & \textbf{16.91} & 56.18 & \underline{40.66} & \underline{43.77} \\ 
\llamaseventy & 8.39 & 4.97 & 6.24 & 15.38 & 9.11 & 11.44 & 21.29 & 12.60 & 15.83 & 58.97 & 37.66 & 42.89 \\ 
\gemmanine & 8.10 & 4.53 & 5.81 & 15.78 & 8.83 & 11.32 & 21.39 & 11.97 & 15.35 & \textbf{61.53} & \textbf{40.76} & \textbf{45.74} \\
\gemmatwentyseven & \textbf{9.84} & \textbf{5.92} & \textbf{7.40} & \underline{15.92} & \underline{9.58} & \underline{11.96} & \underline{21.60} & \underline{13.00} & \underline{16.23} & \underline{60.40} & 37.04 & 42.54 \\
\qwen & 8.67 & 4.41 & 5.85 & 15.85 & 8.07 & 10.70 & 21.00 & 10.70 & 14.17 & 38.56 & 19.64 & 26.03 \\
\deepseek & 3.71 & 1.83 & 2.45 & 7.66 & 3.78 & 5.06 & 9.67 & 4.77 & 6.39 & 53.42 & 32.32 & 36.14 \\
\bottomrule
\end{tabular}
}
\caption{Performance comparison of different models for FOE task  across various evaluation metrics on Reddit type.}
\label{tab:subone_reddit}
\end{table*}

\begin{table*}[!t]
\centering
\small
\resizebox{0.99\linewidth}{!}{
\begin{tabular}{lcccccccccccc}
\toprule
\multirow{2.5}{*}{\textbf{Models}}
& \multicolumn{3}{c}{\textbf{EM}} 
& \multicolumn{3}{c}{\textbf{L-RM}} 
& \multicolumn{3}{c}{\textbf{S-RM}} 
& \multicolumn{3}{c}{\textbf{CM}} \\
\cmidrule(lr){2-4}\cmidrule(lr){5-7}\cmidrule(lr){8-10}\cmidrule(lr){11-13}
& \textbf{Pre} & \textbf{Rec} & \textbf{F1} 
& \textbf{Pre} & \textbf{Rec} & \textbf{F1} 
& \textbf{Pre} & \textbf{Rec} & \textbf{F1} 
& \textbf{Pre} & \textbf{Rec} & \textbf{F1} \\ \midrule
\multicolumn{13}{c}{\textbf{\textit{Proprietary LLMs}}} \\ \midrule
\gptmini & 4.87 & 2.93 & 3.66 & 13.24 & 7.97 & 9.95 & 17.72 & 10.67 & 13.32 & \textbf{77.83} & 48.15 & \underline{57.65} \\ 
\gptfouro & 3.92 & 3.17 & 3.50 & 10.81 & 8.73 & 9.66 & 19.23 & 15.53 & 17.19 & 70.44 & \textbf{57.93} & \textbf{61.80} \\
\claudehaiku & \underline{5.54} & \underline{3.40} & \underline{4.21} & \underline{19.10} & \underline{11.72} & \underline{14.53} & \underline{27.70} & \underline{17.00} & \underline{21.07} & \underline{73.41} & 46.84 & 56.01 \\
\claudesonnet & \textbf{14.01} & \textbf{9.44} & \textbf{11.28} & \textbf{29.16} & \textbf{19.64} & \textbf{23.47} & \textbf{36.64} & \textbf{24.68} & \textbf{29.49} & 70.94 & \underline{49.52} & 57.14 \\ \midrule
\multicolumn{13}{c}{\textbf{\textit{Open-source LLMs}}} \\ \midrule
\llamaeight & \textbf{9.82} & \textbf{7.56} & \textbf{8.54} & \textbf{21.31} & \textbf{16.41} & \textbf{18.54} & \textbf{28.46} & \textbf{21.92} & \textbf{24.77} & \underline{64.15} & 50.72 & 54.89 \\
\llamaseventy & 5.90 & 4.63 & 5.19 & \underline{17.16} & 13.48 & 15.10 & \underline{24.33} & 19.11 & 21.41 & \textbf{66.23} & 52.56 & \textbf{57.13} \\
\gemmanine & \underline{7.24} & 5.22 & \underline{6.06} & 16.99 & 12.25 & 14.24 & 21.87 & 15.77 & 18.32 & 63.11 & 48.66 & 53.49 \\
\gemmatwentyseven & 6.39 & \underline{5.63} & 5.99 & 16.98 & \underline{14.95} & \underline{15.90} & 23.97 & \underline{21.10} & \underline{22.44} & 62.52 & \textbf{53.93} & \underline{56.37} \\
\qwen & 3.55 & 2.99 & 3.24 & 9.74 & 8.21 & 8.91 & 17.66 & 14.89 & 16.16 & 62.82 & \underline{53.48} & 55.82 \\
\deepseek & 2.22 & 1.11 & 1.48 & 5.25 & 2.64 & 3.51 & 10.50 & 5.28 & 7.02 & 63.69 & 34.16 & 42.16 \\
\bottomrule
\end{tabular}
}
\caption{Performance comparison of different models for FOE task  across various evaluation metrics on Review Site type.}
\label{tab:subone_review_site}
\end{table*}

\begin{table*}[!t]
\centering
\small
\resizebox{0.99\linewidth}{!}{
\begin{tabular}{lcccccccccccc}
\toprule
\multirow{2.5}{*}{\textbf{Models}}
& \multicolumn{3}{c}{\textbf{EM}} 
& \multicolumn{3}{c}{\textbf{L-RM}} 
& \multicolumn{3}{c}{\textbf{S-RM}} 
& \multicolumn{3}{c}{\textbf{CM}} \\
\cmidrule(lr){2-4}\cmidrule(lr){5-7}\cmidrule(lr){8-10}\cmidrule(lr){11-13}
& \textbf{Pre} & \textbf{Rec} & \textbf{F1} 
& \textbf{Pre} & \textbf{Rec} & \textbf{F1} 
& \textbf{Pre} & \textbf{Rec} & \textbf{F1} 
& \textbf{Pre} & \textbf{Rec} & \textbf{F1} \\ \midrule
\multicolumn{13}{c}{\textbf{\textit{Proprietary LLMs}}} \\ \midrule
\gptmini & 4.81 & 2.62 & 3.39 & 12.87 & 7.02 & 9.09 & 16.50 & 9.00 & 11.65 & \textbf{63.77} & 42.11 & 45.49 \\
\gptfouro & \underline{10.73} & \textbf{8.51} & \textbf{9.49} & \underline{19.78} & \textbf{15.68} & \textbf{17.49} & \underline{23.33} & \textbf{18.50} & \textbf{20.63} & 61.28 & \textbf{53.61} & \textbf{51.81} \\ 
\claudehaiku & 9.13 & 4.75 & 6.25 & 16.27 & 8.46 & 11.13 & 21.41 & 11.13 & 14.64 & \underline{63.03} & 43.5 & 46.79 \\
\claudesonnet & \textbf{11.61} & \underline{7.62} & \underline{9.20} & \textbf{20.95} & \underline{13.75} & \underline{16.60} & \textbf{25.47} & \underline{16.72} & \underline{20.19} & 60.1 & \underline{48.79} & \underline{48.93} \\ \midrule
\multicolumn{13}{c}{\textbf{\textit{Open-source LLMs}}} \\ \midrule
\llamaeight & \textbf{10.65} & \underline{8.90} & \textbf{9.70} & \underline{17.75} & \underline{14.84} & \underline{16.16} & 21.07 & 17.61 & 19.18 & 48.28 & 49.99 & 44.42 \\
\llamaseventy & 9.93 & \textbf{8.95} & \underline{9.42} & \textbf{17.78} & \textbf{16.02} & \textbf{16.86} & \textbf{21.19} & \textbf{19.09} & \textbf{20.08} & 50.54 & \textbf{51.07} & \textbf{45.76} \\ 
\gemmanine & 6.76 & 5.49 & 6.06 & 13.83 & 11.23 & 12.39 & 15.66 & 12.71 & 14.03 & \underline{52.44} & 48.14 & 45.38 \\
\gemmatwentyseven & 9.45 & 8.36 & 8.87 & 16.22 & 14.34 & 15.22 & 20.64 & \underline{18.25} & \underline{19.37} & 50.8 & \underline{50.02} & \underline{45.46} \\
\qwen & \underline{10.45} & 7.62 & 8.81 & 17.38 & 12.66 & 14.65 & \underline{21.18} & 15.43 & 17.85 & \textbf{52.97} & 45.63 & 43.92 \\
\deepseek & 4.50 & 3.12 & 3.68 & 7.64 & 5.29 & 6.25 & 9.21 & 6.38 & 7.54 & 45.64 & 36.73 & 34.94 \\
\bottomrule
\end{tabular}
}
\caption{Performance comparison of different models for FOE task  across various evaluation metrics on Youtube type.}
\label{tab:subone_youtube}
\end{table*}

%% file: TAB/082abstractive.tex
\begin{table*}[!t]
\setlength{\tabcolsep}{4pt}
\centering
\resizebox{0.90\linewidth}{!}{
\begin{tabular}{l *{5}{c} *{5}{c}}
\toprule
& \multicolumn{5}{c}{\textbf{Blog}} 
& \multicolumn{5}{c}{\textbf{Reddit}} \\
\cmidrule(lr){2-6} \cmidrule(lr){7-11}
\multirow{1}{*}{\textbf{Models}}
& \multicolumn{3}{c}{\textbf{Lexical}} & \multicolumn{2}{c}{\textbf{Semantic}}
& \multicolumn{3}{c}{\textbf{Lexical}} & \multicolumn{2}{c}{\textbf{Semantic}} \\ 
\cmidrule(lr){2-4} \cmidrule(lr){5-6} \cmidrule(lr){7-9} \cmidrule(lr){10-11}
& R-1 & R-2 & R-L & BS & A3CU 
& R-1 & R-2 & R-L & BS & A3CU \\ 
\midrule
\multicolumn{11}{c}{\textbf{\textit{Proprietary LLMs}}} \\ \midrule
\gptmini           & \textbf{40.14} & \textbf{16.32} & \textbf{36.06} & \textbf{90.32} & \textbf{36.03}
                    & \underline{38.41} & \underline{12.60} & \underline{33.15} & \underline{90.17} & \underline{37.71} \\
\gptfouro          & \underline{37.06} & \underline{14.68} & \underline{33.03} & \underline{89.18} & \underline{31.34}
                    & \textbf{41.13} & \textbf{15.33} & \textbf{36.15} & \textbf{90.22} & \textbf{40.65} \\
\claudehaiku       & 31.15 & 9.91  & 27.01 & 87.45 & 25.75
                    & 35.20 & 10.49 & 30.42 & 89.10 & 34.22 \\ 
\claudesonnet      & 31.87 & 9.04  & 28.44 & 87.96 & 25.69
                    & 34.77 & 9.88  & 30.49 & 89.28 & 34.62 \\ 
\midrule
\multicolumn{11}{c}{\textbf{\textit{Open-source LLMs}}} \\ \midrule
\llamaeight        & 35.77 & \underline{14.10} & 32.63 & 89.52 & 30.27
                    & 33.73 & 9.26  & 29.58 & 85.82 & 28.52 \\ 
\llamaseventy      & \textbf{38.82} & \textbf{15.65} & \textbf{34.70} & \textbf{90.09} & \textbf{32.22}
                    & \textbf{37.60} & \textbf{12.13} & \textbf{32.76} & \underline{89.70} & \textbf{35.24} \\ 
\gemmanine         & \underline{35.86} & 13.55 & \underline{33.45} & 89.77 & 28.95
                    & 35.00 & 10.31 & 30.54 & 88.36 & 30.72 \\
\gemmatwentyseven  & 34.73 & 12.42 & 31.07 & 89.74 & \underline{31.00}
                    & \underline{36.27} & \underline{11.04} & \underline{31.38} & \textbf{90.01} & \underline{34.36} \\ 
\qwen              & 35.54 & 12.45 & 32.29 & \underline{89.88} & 26.30
                    & 31.27 & 8.74  & 26.92 & 89.37 & 25.24 \\ 
\deepseek          & 34.79 & 11.75 & 30.95 & 73.04 & 21.09
                    & 35.21 & 10.22 & 30.89 & 75.50 & 25.98 \\ 
\bottomrule
\end{tabular}
}
\caption{Performance comparison of different models for the OIG task using lexical and semantic metrics on the Blog and Reddit Types.}
\label{tab:oig_blog_reddit}
\end{table*}

\begin{table*}[!t]
\setlength{\tabcolsep}{4pt}
\centering
\resizebox{0.90\linewidth}{!}{
\begin{tabular}{l *{5}{c} *{5}{c}}
\toprule
& \multicolumn{5}{c}{\textbf{Review Site}} 
& \multicolumn{5}{c}{\textbf{YouTube}} \\
\cmidrule(lr){2-6} \cmidrule(lr){7-11}
\multirow{1}{*}{\textbf{Models}}
& \multicolumn{3}{c}{\textbf{Lexical}} & \multicolumn{2}{c}{\textbf{Semantic}}
& \multicolumn{3}{c}{\textbf{Lexical}} & \multicolumn{2}{c}{\textbf{Semantic}} \\ 
\cmidrule(lr){2-4} \cmidrule(lr){5-6} \cmidrule(lr){7-9} \cmidrule(lr){10-11}
& R-1 & R-2 & R-L & BS & A3CU 
& R-1 & R-2 & R-L & BS & A3CU \\ 
\midrule
\multicolumn{11}{c}{\textbf{\textit{Proprietary LLMs}}} \\ \midrule
\gptmini           & \textbf{39.85} & \underline{14.35} & \textbf{35.24} & \textbf{90.42} & \textbf{41.60}
                    & \underline{38.90} & \underline{12.82} & \underline{33.90} & \textbf{90.47} & \underline{38.96} \\
\gptfouro          & \underline{39.44} & \textbf{14.50} & \underline{34.98} & \underline{89.83} & \underline{41.07}
                    & \textbf{40.15} & \textbf{14.67} & \textbf{35.65} & \underline{90.28} & \textbf{41.45} \\
\claudehaiku       & 34.57 & 10.20 & 29.76 & 88.70 & 36.87
                    & 33.36 & 9.76  & 29.14 & 88.76 & 31.57 \\ 
\claudesonnet      & 33.80 & 9.44  & 29.22 & 88.84 & 34.34
                    & 34.12 & 9.57  & 30.03 & 89.11 & 32.47 \\ 
\midrule
\multicolumn{11}{c}{\textbf{\textit{Open-source LLMs}}} \\ \midrule
\llamaeight        & \textbf{40.97} & \textbf{15.85} & \textbf{36.60} & 90.05 & \underline{38.24}
                    & 34.77 & 9.88  & 30.49 & 89.66 & \underline{33.89} \\ 
\llamaseventy      & \underline{36.00} & 12.00 & \underline{31.64} & \underline{90.34} & 37.30
                    & \textbf{44.88} & \textbf{17.38} & 29.49 & \textbf{90.64} & \textbf{36.70} \\ 
\gemmanine         & 34.10 & 11.17 & 30.06 & 88.45 & 35.43
                    & 34.84 & 10.54 & 30.70 & 86.36 & 29.41 \\ 
\gemmatwentyseven  & 35.48 & \underline{12.43} & 30.84 & \textbf{90.46} & \textbf{38.58}
                    & 35.12 & 10.88 & \underline{30.87} & \underline{90.10} & 32.90 \\ 
\qwen              & 33.10 & 10.71 & 29.61 & 89.73 & 25.29
                    & 35.37 & \underline{11.27} & 22.89 & 89.27 & 24.47 \\ 
\deepseek          & 33.74 & 9.81  & 29.01 & 83.65 & 29.12
                    & \underline{36.47} & 10.84 & \textbf{32.14} & 77.44 & 27.21 \\ 
\bottomrule
\end{tabular}
}
\caption{Performance comparison of different models for the OIG task using lexical and semantic metrics on the Review Site and YouTube Types.}
\label{tab:oig_review_youtube}
\end{table*}

%% file: TAB/083prompt.tex
\begin{table*}
    \small
    \centering
    \begin{tabular}{p{14cm}}
    \toprule
    \textbf{\subone (FOE) task prompt} \\
    \midrule
\textcolor{teal}{\textbf{[Task Description]}}\\
You are a car opinion miner for the user. 
Your task is to extract tuples in the form of (entity, feature, opinion) by identifying attributes, specific features, or components mentioned in the text and associating opinions with each feature.
\\\\
1. A "entity" is the name of the car model with brand which opinions are expressed (e.g., "volvo xc90", "toyota camry", "Nissan Sentra").
\\2. A "feature" as a specific characteristic, attribute, or component of an entity that users mention or evaluate (e.g., "interior design," "fuel efficiency," "safety features").
   - The performance, design, or experience (e.g., "handling", "ride comfort").
   - Distinct functions or technologies in a vehicle (e.g., "infotainment system").
   - Physical parts or systems that make up the vehicle (e.g., "brake", "transmission").
\\3. An "opinion" is a subjective or objective judgment, reaction, experience, evaluation, or feedback about the entity's feature, including assessments of quality, performance, or value, as well as direct responses or reactions from users based on their experience.
\\\\

\textbf{IMPORTANT}:

\begin{itemize}[leftmargin=*,topsep=4pt,itemsep=4pt,parsep=0pt]
\item Features and opinions MUST be extracted in the input text. Never generate words or terms that do not exist in the text.
\item However, if the feature corresponding to an opinion does not exist in the text and is implicit, the feature is treated as "NULL".
\item The opinion MUST be no more than 5 words.
\item The output must be in valid JSON format, but **DO NOT** include "json" code block delimiters (e.g., ```json ... ```). 
\item Return **only** the JSON object, without any extra text, explanations, or comments.
\item Provide only the tuples. Do not mention your process or how you arrived at it.
\item Note: Return your results in JSON format only, with the following structure:
\texttt{\{'opinion\_tuple': [\{'entity': <str>, 'feature': <str>, 'opinion': <str>\}, \{'entity': <str>, 'feature': <str>, 'opinion': <str>\}, ..., \{'entity': <str>, 'feature': <str>, 'opinion': <str>\}]\}}
\end{itemize}
\\
\textcolor{teal}{\textbf{[Content Text]}}\\
...
\\
\bottomrule
    \end{tabular}
    \caption{The prompt for \subone.}
    \label{tab:subone_prompt}
\end{table*}

\begin{table*}
    \small
    \centering
    \begin{tabular}{p{14cm}}
    \toprule
    \textbf{\subtwo (OIG) task prompt} \\
    \midrule
\textcolor{teal}{\textbf{[Task Description]}}\\
You are a product and marketing manager at a global automotive company. Your task is to produce a free-form insight text that categorizes and organizes a user text into higher-level insights, such that the report alone provides a clear understanding of the key opinions expressed. This insight should be written in natural, human-like language and structured around the core topics (features).
\\\\
\textbf{Step to Follow}:
\\
\textbf{1. Read and Understand}
\begin{itemize}[leftmargin=*,topsep=4pt,itemsep=4pt,parsep=0pt]
    \item Examine the online text to identify its main points.
\end{itemize}
\\
\textbf{2. Organize Top-Level Topics}
\begin{itemize}[leftmargin=*,topsep=4pt,itemsep=4pt,parsep=0pt]
    \item Group similar or related tuples into clear categories (e.g., “Engine Issues,” “Warranty Feedback,” etc.).
    \item Reflect on the frequency or intensity of opinions if it helps convey importance.
\end{itemize}
\\
\textbf{3. Create a Three-Line Report}
\begin{itemize}[leftmargin=*,topsep=4pt,itemsep=4pt,parsep=0pt]
    \item Line 1: Highlight the most frequently mentioned or emphasized features, grouping related opinions into high-level categories.
    \item Line 2: Focus on the features that a user strongly praised or criticized, incorporating the intensity or frequency of opinions where applicable.
    \item Line 3: Provide a cohesive conclusion summarizing the overarching sentiment or key takeaway from a user’s text.
\end{itemize}
\\
\textbf{4. Write the Insight}
\begin{itemize}[leftmargin=*,topsep=4pt,itemsep=4pt,parsep=0pt]
    \item Make it short (three to five lines).
    \item Use clear, direct language.
    \item Ensure that reading only this insight sufficiently conveys a user’s main viewpoints.
\end{itemize}
\\
\textbf{What to Avoid:}
\begin{itemize}[leftmargin=*,topsep=4pt,itemsep=4pt,parsep=0pt]
    \item Provide only the insight text. Do not mention your process or how you arrived at it.
    \item Do not include introductory phrases such as “Here is a insight of the review” or “Based on the review text.”
    \item Do not directly copy sentences from the online text; rephrase and synthesize information.
    \item Carefully analyze the given text to determine the number of users and decide whether to use "user" or "users" accordingly.
\end{itemize}
Note: Return your results in JSON format only, with the following structure: \texttt{\{'insight': <str>\}}
\\\\
\textcolor{teal}{\textbf{[Content Text]}}
\\
...
\\
\bottomrule
    \end{tabular}
    \caption{The prompt for \subtwo.}
    \label{tab:subtwo_prompt}
\end{table*}

\begin{table*}
\small
\centering
\begin{tabular}{p{14cm}}
\toprule
\textbf{Contextual Match (CM) prompt} \\
\midrule
\textcolor{teal}{\textbf{[Task Description]}}  
\\You are given two lists of tuples, each in the form [(entity, feature, opinion), (entity, feature, opinion), ...].  
One list represents the Gold (correct) tuples, and the other list represents the Model's Predicted tuples.  
Your goal is to calculate four values:  
- matched\_pred\_tuple: The matched Predicted tuple.  
- matched\_gold\_tuple: The matched Gold tuple.  

\\\textbf{Match Criteria}:
\\- Examine the two lists of tuples to identify their main points.  
\\- Convert each element (entity, feature, opinion) to lowercase before comparing (e.g., "kia soul" vs. "Kia Soul" are equivalent).  
\\- Allow flexibility when matching tuples by considering semantic equivalence, synonyms, rephrased expressions, or other valid variations that convey the same context or meaning. For instance, the following cases should be considered as valid matches:  
\begin{itemize}[leftmargin=*,topsep=4pt,itemsep=4pt,parsep=0pt]
  \item Pred: ("toyota corolla", "brakes", "getting hot"), Gold: ("toyota corolla le", "brakes", "getting hot")  
  \item Pred: ("toyota camry", "null", "looks better"), Gold: ("camry", "looks", "better")  
  \item Pred: ("porsche 911", "performance", "can go effortlessly fast"), Gold: ("porsche 911", "drive", "effortlessly fast")  
  \item Pred: ("aston martin vanquish", "rear badge", "would look better"), Gold: ("aston martin vanquish", "badge", "better")  
\end{itemize}
- A tuple is considered a relaxed match if all three elements are semantically equivalent after applying these transformations.  
\\- Do not count a match more than once if there are duplicates.  
\\ - Note: Return your results in JSON format only, with the following structure: 
\\ \texttt{\{\textquotesingle matched\_tuple\_pair\textquotesingle: [
\{\textquotesingle matched\_pred\_tuple\textquotesingle: (\textquotesingle entity\textquotesingle, \textquotesingle feature\textquotesingle, \textquotesingle opinion\textquotesingle),
\textquotesingle matched\_gold\_tuple\textquotesingle: (\textquotesingle entity\textquotesingle, \textquotesingle feature\textquotesingle, \textquotesingle opinion\textquotesingle) ... \}}, 
\\\\ \textcolor{teal}{\textbf{[Pred Tuples]}}\\
...
\\\\ \textcolor{teal}{\textbf{[Gold Tuples]}}
\\ ...\\ 
\bottomrule
\end{tabular}
\caption{The prompt for Contextual Match (CM).}
\label{tab:gm_prompt}
\end{table*}

\begin{table*}
\small
\centering
\begin{tabular}{p{14cm}}
\toprule
\textbf{LLM-as-a-judge Evaluation prompt - (Faithfulness) } \\
\midrule
\textcolor{teal}{\textbf{[Task Description]}}  
\\You are provided with an online user's automobile review text along with an opinion-centric insight summary that groups user opinions at the topic level to offer insights. Your task is to evaluate the insight based on its Faithfulness.
Make sure you understand the following evaluation metric very clearly.
\\\\Evaluation Criteria:

Faithfulness (1-5) – Evaluate whether the insight accurately represents the original review without distorting its meaning, omitting key details, or introducing hallucinated information that was not present in the original text.
\\• Score 1: The insight completely distorts the original online content \\and contains many incorrect information. It cannot be trusted at all.
\\• Score 2: The insight significantly misrepresents the original online content with several incorrect information.
\\• Score 3: The insight partially reflects the original online content but has some minor incorrect information.
\\• Score 4: The insight significantly misrepresents the original online content with several incorrect information.
\\• Score 5: The insight completely reflects the original online content without any distortions and incorrect information.
\\\\\textcolor{teal}{\textbf{[Review Text:]}}  
...

\\\textcolor{teal}{\textbf{[Insight:]}}  
...
\\\\Evaluation Steps:
\\1. Read through the review text provided.
\\2. Read the corresponding opinion-centric insight summary.
\\3. Evaluate whether the insight faithfully reflects the original review without any distortion.
\\4. Check if the insight contains any hallucinated information not present in the review.
\\5. Assign a score from 1 to 5 for Faithfulness, where 1 indicates very poor faithfulness and 5 indicates excellent faithfulness.
\\ 
\bottomrule
\end{tabular}
\caption{The prompt for LLM-as-a-judge Evaluation (Faithfulness).}
\label{tab:llm_judge_faithfulness}
\end{table*}

\begin{table*}
\small
\centering
\begin{tabular}{p{14cm}}
\toprule
\textbf{LLM-as-a-judge Evaluation prompt - (Coverage) } \\
\midrule
\textcolor{teal}{\textbf{[Task Description]}}  
\\You are provided with an online user's automobile review text along with an opinion-centric insight summary that groups user opinions at the topic level to offer insights. Your task is to evaluate the insight based on its Coverage.
Make sure you understand the following evaluation metric very clearly.
\\\\Evaluation Criteria:
Coverage (1-5) – Evaluate whether the insight effectively captures and represents the key opinions expressed in the review without omitting significant details or key points. Ensure that all essential opinions are included and accurately represented.
\\•Score 1: The insight fails to capture any key opinions from the online content. The content is either completely missing or irrelevant to the original opinions.
\\•Score 2: The insight captures only a small portion of key opinions. Many important opinions from the online content are missing.
\\•Score 3: The insight captures some key opinions but misses others. The coverage is partial and could be more comprehensive.
\\•Score 4: The insight effectively captures most key opinions from the online content. The coverage is good but may miss minor details.
\\•Score 5: The insight comprehensively captures all key opinions from the online content. Nothing important is missing, and the coverage is complete.
\\\\\textcolor{teal}{\textbf{[Review Text:]}}  
...

\\\textcolor{teal}{\textbf{[Insight:]}}  
...
\\\\Evaluation Steps:
\\1. Read through the review text provided.
\\2. Read the corresponding opinion-centric insight summary.
\\3. Evaluate whether the insight effectively captures and represents the key opinions expressed in the review.
\\4. Consider if any significant details or opinions are missing from the insight.
\\5. Assign a score from 1 to 5 for Coverage, where 1 indicates very poor coverage and 5 indicates excellent coverage.
\\ 
\bottomrule
\end{tabular}
\caption{The prompt for LLM-as-a-judge Evaluation (Coverage).}
\label{tab:llm_judge_coverage}
\end{table*}

\begin{table*}
\small
\centering
\begin{tabular}{p{14cm}}
\toprule
\textbf{LLM-as-a-judge Evaluation prompt - (Specificity) } \\
\midrule
\textcolor{teal}{\textbf{[Task Description]}}  
\\You are provided with an online user's automobile review text along with an opinion-centric insight insight that groups user opinions at the topic level to offer insights. Your task is to evaluate the insight based on its Specificity.
Make sure you understand the following evaluation metric very clearly.
\\\\Evaluation Criteria:
Specificity (1-5) – Evaluate whether the insight presents meaningful and relevant details by including concrete information rather than being vague or overly generic. Ensure that the insight provides clear, detailed points that directly relate to the review content.
\\•Score 1: The insight is extremely vague and generic, lacking any meaningful details. It uses broad generalizations without specific examples or descriptions.
\\•Score 2: The insight includes very few specific details. Most information is presented in a general way without concrete examples.
\\•Score 3: The insight includes some specific details but could be more precise. There is a mix of specific and generic information.
\\•Score 4: The insight provides good specific details in most areas. The information is concrete and meaningful, though some minor points could be more detailed.
\\•Score 5: The insight is highly specific throughout, providing precise and meaningful details. All information is concrete with relevant examples and descriptions.
\\\\\textcolor{teal}{\textbf{[Review Text:]}}  
...

\\\textcolor{teal}{\textbf{[Insight:]}}  
...
\\\\Evaluation Steps:
\\1. Read through the review text provided.
\\2. Read the corresponding opinion-centric insight summary.
\\3. Evaluate whether the insight provides specific, concrete details and avoids overly general or ambiguous statements.
\\4. Assign a score from 1 to 5 for Specificity, where 1 indicates very poor specificity and 5 indicates excellent specificity.
\\ 
\bottomrule
\end{tabular}
\caption{The prompt for LLM-as-a-judge Evaluation (Specificity).}
\label{tab:llm_judge_specificity}
\end{table*}

\begin{table*}
\small
\centering
\begin{tabular}{p{14cm}}
\toprule
\textbf{LLM-as-a-judge Evaluation prompt - (Insightfulness) } \\
\midrule
\textcolor{teal}{\textbf{[Task Description]}}  
\\You are provided with an online user's automobile review text along with an opinion-centric insight summary that groups user opinions at the topic level to offer insights. Your task is to evaluate the insight based on its insightfulness. Make sure you understand the following evaluation metric very clearly.
\\\\Evaluation Criteria:
Insightfulness (1-5) – Evaluate whether the insight provides meaningful insights that enhance understanding or decision-making for the reader. Ensure that the insight goes beyond a simple restatement of the review to offer unique interpretations or conclusions that add value.
\\•Score 1: The insight provides no meaningful insights. It simply restates basic facts without adding any value for understanding or decision-making.
\\•Score 2: The insight offers very limited insights. Most information is superficial and does not help readers gain deeper understanding.
\\•Score 3: The insight provides some useful insights but could go deeper. It offers moderate value for understanding and decision-making.
\\•Score 4: The insight provides good insights in most areas. The information is valuable for understanding and decision-making, though some points could be more insightful.
\\•Score 5: The insight provides excellent insights throughout. All information meaningfully enhances understanding and is highly valuable for decision-making.
\\\\\textcolor{teal}{\textbf{[Review Text:]}}  
...

\\\textcolor{teal}{\textbf{[Insight:]}}  
...
\\\\Evaluation Steps:
\\1. Read through the review text provided.
\\2. Read the corresponding opinion-centric insight summary.
\\3. Evaluate whether the insight offers insightful, value-added interpretations that deepen understanding or guide decisions.
\\4. Assign a score from 1 to 5 for Insightfulness., where 1 indicates very poor Insightfulness. and 5 indicates excellent insightfulness.
\\ 
\bottomrule
\end{tabular}
\caption{The prompt for LLM-as-a-judge Evaluation (Insightfulness).}
\label{tab:llm_judge_insightfulness}
\end{table*}

\begin{table*}
\small
\centering
\begin{tabular}{p{14cm}}
\toprule
\textbf{LLM-as-a-judge Evaluation prompt - (Intent) } \\
\midrule
\textcolor{teal}{\textbf{[Task Description]}}  
\\You are provided with an online user's automobile review text along with an opinion-centric insight summary that groups user opinions at the topic level to offer insights. Your task is to evaluate the insight based on its Intent.
Make sure you understand the following evaluation metric very clearly.
\\\\Evaluation Criteria:
Intent (1-5) – Evaluate whether the insight accurately preserves the author's original tone, intent, and nuances without altering the emotional or stylistic essence of the review. Consider if the insight maintains the original sentiment and communication style throughout.
\\•Score 1: The insight completely fails to preserve the original tone and intent. The emotional essence and nuances are lost or significantly distorted.
\\•Score 2: The insight largely misrepresents the original tone and intent. Many nuances are missed or altered, though some basic sentiments remain intact.
\\•Score 3: The insight somewhat preserves the original tone and intent. Some nuances are captured while others are missed or altered.
\\•Score 4: The insight generally preserves the original tone and intent well. Most nuances and emotional elements are accurately captured, with only minor alterations.
\\•Score 5: The insight perfectly preserves the original tone, intent, and nuances. The emotional and stylistic essence is captured with complete accuracy.
\\\\\textcolor{teal}{\textbf{[Review Text:]}}  
...

\\\textcolor{teal}{\textbf{[Insight:]}}  
...
\\\\Evaluation Steps:
\\1. Read through the review text provided.
\\2. Read the corresponding opinion-centric insight summary.
\\3. Evaluate whether the insight maintains the original tone, intent, and nuanced details of the review without altering its emotional or stylistic qualities.
\\4. Assign a score from 1 to 5 for Intent, where 1 indicates very poor preservation and 5 indicates excellent preservation.
\\ 
\bottomrule
\end{tabular}
\caption{The prompt for LLM-as-a-judge Evaluation (Intent).}
\label{tab:llm_judge_intent}
\end{table*}

\begin{table*}
\small
\centering
\begin{tabular}{p{14cm}}
\toprule
\textbf{LLM-as-a-judge Evaluation prompt - (Fluency) } \\
\midrule
\textcolor{teal}{\textbf{[Task Description]}}  
\\You are provided with an online user's automobile review text along with an opinion-centric insight summary that groups user opinions at the topic level to offer insights. Your task is to evaluate the insight based on its Fluency.
Make sure you understand the following evaluation metric very clearly.
\\\\Evaluation Criteria:
Fluency (1-5) – Evaluate whether the insight is naturally written, grammatically correct, and easy to read. Consider whether the insight flows smoothly, uses proper grammar, and avoids awkward phrasing, ensuring it is accessible to the reader.
\\•Score 1: The insight is extremely difficult to read. It is filled with errors, awkward phrasing, and lacks proper grammar.
\\•Score 2: The insight is difficult to read. It contains many errors, awkward phrasing, and lacks proper grammar.
\\•Score 3: The insight is somewhat difficult to read. It has some errors, awkward phrasing, and lacks proper grammar.
\\•Score 4: The insight is generally easy to read. It has few errors, awkward phrasing, and lacks proper grammar.
\\•Score 5: The insight is extremely easy to read. It is filled with errors, awkward phrasing, and lacks proper grammar.
\\\\\textcolor{teal}{\textbf{[Review Text:]}}  
...

\\\textcolor{teal}{\textbf{[Insight:]}}  
...
\\\\Evaluation Steps:
\\1. Read through the review text provided.
\\2. Read the corresponding opinion-centric insight summary.
\\3. Evaluate whether the insight is written in a natural and grammatically correct manner with clear and smooth flow.
\\4. Assign a score from 1 to 5 for Fluency, where 1 indicates very poor fluency and 5 indicates excellent fluency.
\\ 
\bottomrule    
\end{tabular}
\caption{The prompt for LLM-as-a-judge Evaluation (Fluency).}
\label{tab:llm_judge_fluency}
\end{table*}

%% file: TAB/087annotatoin_guidlines.tex
\begin{table*}
\small
\centering
\begin{tabular}{p{14cm}}
\toprule
\textbf{Entity-feature-opinion tuple Annotation Prompt} \\
\midrule
\textcolor{teal}{\textbf{[Task Definition]}}
\\
You are a car opinion miner for the user. I want to perform feature-centric opinion extraction which is identifying attributes, specific features, or components mentioned in the text and associating opinions with each feature. To maximize coverage and diversity, you will generate a comprehensive set of tuples using your reasoning and understanding of the text. You will receive a user-generated car-related text. Your task is to extract all possible tuples in the form of (entity, feature, opinion) that exist within the text.

\begin{itemize}[leftmargin=*,topsep=4pt,itemsep=4pt,parsep=0pt]
    \item \textbf{Entity}: the brand and model of a vehicle for which an opinion is expressed (e.g., "volvo xc90", "toyota camry", "nissan sentra").
    \item \textbf{Feature}: Feature: a specific characteristic, attribute, or component of the entity that users mention or evaluate. Includes aspects of performance, design, driving experience, features, systems, or parts (e.g., "ride comfort", "handling", "infotainment system", "interior", "brakes").
    \item \textbf{Opinion}: Opinion: a subjective or objective evaluation, reaction, or judgment about a feature. The opinion span must be taken exactly from the text and contain no more than 5 words. If the feature is implicit (not explicitly stated in the text), label it as "NULL".
    \item \textbf{Evidence Sentence}: the exact sentence from the input text that contains both the feature and the opinion (or where the opinion is clearly expressed about the implicit feature).
\end{itemize}

\\
\textcolor{teal}{\textbf{[Example]}}
\begin{itemize}[leftmargin=*,topsep=4pt,itemsep=4pt,parsep=0pt]
    
    \item \textbf{Text}: The EyeSight system of Toyota Camry SE is amazing compared to Toyotas whatever it is.
    \item \textbf{Entity}: Toyota Camry SE
    \item \textbf{Feature}: EyeSight system
    \item \textbf{Opinion}: amazing compared to Toyotas whatever it is
\end{itemize}
\\
\textcolor{teal}{\textbf{[Annotation Requirements]}}
\begin{itemize}[leftmargin=*,topsep=4pt,itemsep=4pt,parsep=0pt]
    \item All tuples must come only from the input text. Never invent or infer content that does not exist in the text.
    \item Do not hallucinate any information or generate tuples not explicitly supported by the input.
    \item Do not include any explanations, reasoning, or formatting outside the JSON object.
    \item The output must be in valid JSON format, and contain only the JSON object as follows:
    \texttt{\{'opinion\_tuple': [\{'entity': <str>, 'feature': <str>, 'opinion': <str>, 'evidence': <str>\}, \{'entity': <str>, 'feature': <str>, 'opinion': <str>, 'evidence': <str>\}, ..., \{'entity': <str>, 'feature': <str>, 'opinion': <str>, 'evidence': <str>\}]\}}
\end{itemize}
\\
\textcolor{teal}{\textbf{[Content Text]}}
\\ ...\\ 
\bottomrule
    \end{tabular}
    \caption{\proposed Entity-feature-opinion tuple annotation prompt.}
    \label{tab:tuple_annotation_prompt}
\end{table*}

\begin{table*}
\small
\centering
\begin{tabular}{p{14cm}}
\toprule
\textbf{Entity-feature-opinion tuple Annotation \& Verification Guideline} \\
\midrule
\textcolor{teal}{\textbf{[Task Definition]}}
\\
Your task is to analyze what kind of opinions users express about cars in online user-generated content.
\begin{itemize}[leftmargin=*,topsep=4pt,itemsep=4pt,parsep=0pt]
    \item \textbf{Entity}: Car brand \& model (e.g., volvo xc90, toyota camry, Nissan Sentra)
    \item \textbf{Feature}: A characteristic, attribute, or component of the mentioned entity (e.g., handling, ride comfort, infotainment system, ...)
    \item \textbf{Opinion}: The user's subjective/objective judgment, reaction, experience, evaluation, or feedback about the feature (e.g., terrible, incredibly stable)
\end{itemize}
\\
\textcolor{teal}{\textbf{[Example]}}
\begin{itemize}[leftmargin=*,topsep=4pt,itemsep=4pt,parsep=0pt]
    
    \item \textbf{Text}: The EyeSight system of Toyota Camry SE is amazing compared to Toyotas whatever it is.
    \item \textbf{Entity}: Toyota Camry SE
    \item \textbf{Feature}: EyeSight system
    \item \textbf{Opinion}: amazing compared to Toyotas whatever it is
\end{itemize}
\\
\textcolor{teal}{\textbf{[Description]}}
\begin{itemize}[leftmargin=*,topsep=4pt,itemsep=4pt,parsep=0pt]
    \item Each web page displays a online content along with a single (entity, feature, opinion) tuple at a time.
    \item A content can contain multiple \textbf{(entity, feature, opinion)} tuples. 
    \item Sentence refers to the evidence sentence for the feature and opinion. 
    \item \textbf{feature\_exist, opinion\_exist}: Whether the feature or opinion exists in the sentence.
    \item \textbf{sentence\_exist}: Whether the sentence exists in the document on the right.
    \item If a feature appears in the content, it is highlighted in \textbf{\textcolor{red}{bold red}} in both the Document Info and Document Text.
    \item If an opinion appears in the document, it is highlighted in \textbf{\textcolor{blue}{bold blue}}.
    \item The evidence sentence is highlighted in \textbf{bold black}.
    \item Each tuple is shown in the \textit{Data Information} section, while the content text appears in the \textit{Document Text} section. Annotators follow the process described below to perform accurate verification and refinement.
\end{itemize}

\\
\textcolor{teal}{\textbf{[Annotation \& Verification Process]}}
\\
\textbf{(1) Entity Verification}
Check if the car brand and name (displayed in the entity field) is correctly identified from the document. The entity typically appears in titles, subtitles, or once in the main text. Confirm the context is indeed about this vehicle and the name is recorded exactly as it appears in the document. 
\\
\\\textbf{(2) Opinion Existence Check}
Verify that opinions exist explicitly in the document as words or phrases. When an opinion is found in the text (highlighted in blue), confirm opinion\_exist is marked 'O'. If no explicit opinion is found in any sentence, verify opinion\_exist is marked 'X'. 
\\
\\\textbf{(3) Feature-Opinion Relationship Verification} 
Check if the feature (highlighted in red) being discussed in relation to the opinion exists explicitly in the document. For explicit features, verify feature\_exist is marked 'O' and the exact text from the document is used. For implicit features (not explicitly mentioned), verify they are marked as 'NULL' but feature\_exist is still 'O'. 
\\
\\\textbf{(4) Sentence Documentation Check}
Verify that the evidence sentence (in bold black) properly contains either both feature and opinion (when they appear in the same sentence) or spans from feature to opinion (when they appear in different sentences). Confirm sentence\_exist is marked 'O' when this evidence appears in the document. 
\\
\\\textbf{(5) Duplicate Tuple Check}
When multiple identical or similar feature-opinion pairs exist in the same document, verify that only one correct sample is kept (with proper feature\_exist and opinion\_exist marks) while others are marked 'X' to avoid duplication. 
\\
\bottomrule
    \end{tabular}
    \caption{\proposed Entity-feature-opinion tuple annotation and verification guideline.}
    \label{tab:tuple_guideline}
\end{table*}


\begin{table*}
\small
\centering
\begin{tabular}{p{14cm}}
\toprule
\textbf{Opinion-centric-insight Annotation Prmopt} \\
\midrule
\textcolor{teal}{\textbf{[Task Definition]}}
\\
You are a product and marketing manager at a global automotive company. You will be given a piece of user-generated automotive content, along with its final gold-standard set of (entity, feature, opinion) tuples. Your goal is to write a free-form opinion-centric insight summary that organizes and synthesizes the content into high-level, insightful categories. 
\\
Here is definition of (entity, feature, opinion) tuple and example:
\begin{itemize}[leftmargin=*,topsep=4pt,itemsep=4pt,parsep=0pt]
    \item \textbf{Entity}: the brand and model of a vehicle for which an opinion is expressed (e.g., "volvo xc90", "toyota camry", "nissan sentra").
    \item \textbf{Feature}: Feature: a specific characteristic, attribute, or component of the entity that users mention or evaluate. Includes aspects of performance, design, driving experience, features, systems, or parts (e.g., "ride comfort", "handling", "infotainment system", "interior", "brakes").
    \item \textbf{Opinion}: Opinion: a subjective or objective evaluation, reaction, or judgment about a feature. The opinion span must be taken exactly from the text and contain no more than 5 words. If the feature is implicit (not explicitly stated in the text), label it as "NULL".
\end{itemize}
This insight should be written in natural language, structured around the core features discussed, and clearly convey the most salient and recurring opinions.
\\
\textbf{Step 1. Read and Understand}

\begin{itemize}[leftmargin=*,topsep=4pt,itemsep=4pt,parsep=0pt]
    \item Thoroughly examine both the input content and the associated tuples.
    \item Identify the main opinion clusters and the intensity or frequency of key topics.
\end{itemize}

\textbf{Step 2. Organize into High-Level Topics}
\begin{itemize}[leftmargin=*,topsep=4pt,itemsep=4pt,parsep=0pt]
    \item Group similar tuples into broad categories (e.g., "Interior Design", "Performance \& Handling").
    \item Reflect on user sentiment and frequency to prioritize key points.
\end{itemize}

\textbf{Step 3. Generate a Three-to-Five-Line Insight}

\begin{itemize}[leftmargin=*,topsep=4pt,itemsep=4pt,parsep=0pt]
    \item Line 1: Highlight the most frequently mentioned or emphasized features.
    \item Line 2: Emphasize features that were praised or criticized with notable intensity or consensus.
    \item Line 3+: Provide a concluding sentence summarizing the overall sentiment or takeaway.
\end{itemize}

\textbf{Important Instructions}
\begin{itemize}[leftmargin=*,topsep=4pt,itemsep=4pt,parsep=0pt]
    \item The insight should be concise, natural, and informative—suitable for a marketing manager’s report
    \item Ensure it covers the major opinions expressed in the tuple set
    \item The tone should be neutral and professional, without exaggeration
    \item Do not directly copy sentences from the content
    \item Do not mention the annotation process or describe how the insight was generated
    \item Do not include phrases like “Based on the review…” or “Here is a insight…”
    \item Carefully analyze whether the input reflects a single user or multiple users, and adjust nouns/pronouns accordingly ("user" vs. "users").
    \item Note: Return your results in JSON format only, with the following structure: \texttt{\{'insight': <str>\}} 
\end{itemize}
\\
\textcolor{teal}{\textbf{[Content Text]}}
\\ ...\\ 
\bottomrule
\end{tabular}
\caption{\proposed: Opinion-centric insight annotation prompt.}
\label{tab:summary_annotation_prompt}
\end{table*}

\begin{table*}
\small
\centering
\begin{tabular}{p{14cm}}
\toprule
\textbf{Opinion-centric-insight Annotation \& Verification Guideline} \\
\midrule
\begin{minipage}[t]{\linewidth} 
\textcolor{teal}{\textbf{[Task Definition]}}
\\
Your task is to analyze a text that insight summarizes the main topics and insights derived from online user-generated content. The provided summary is not a simple condensation, but an opinion-centric insight that synthesizes multiple users’ perspectives to deliver high-level insights useful for marketing and strategic decision-making.
\\\\
\textcolor{teal}{\textbf{[Description]}}
\begin{itemize}[leftmargin=*,topsep=4pt,itemsep=4pt,parsep=0pt]
    \item Each page displays a online content, a single gold tuple from the content, and its corresponding insight.
    \item Each insight is paired with a single gold tuple from the content and displayed on the page. When the page is turned, the next gold tuple appears.
    \item Each insight and a single gold tuple is shown in the \textit{Data Information} section, while the content text appears in the \textit{Document Text} section. Annotators follow the process described below to perform accurate verification and refinement.
\end{itemize}

\textcolor{teal}{\textbf{[Annotation \& Verification Process]}}

\textbf{(1) Read \& Understand Original Text and Insight} 
\\
Carefully read the original user-generated content and Insight. 
\\\\
\textbf{(2) Check Factual Accuracy} 
\\
Thoroughly examine the insight for any hallucinations or factual inaccuracies that contradict information in the original content. Check whether all claims in the insight are directly supported by explicit statements in the source texts. Verify that no fabricated or assumed information is included, particularly for technical specifications, features, or entity attributes mentioned in the insight.
\\\\
\textbf{(3) Check Subject Coherence} 
\\
Verify the correct attribution of opinions to appropriate subjects, considering the multi-user discussion context. Ensure opinions are not misattributed across different entities or users, especially in complex, multi-threaded discussions. Check that the insight properly distinguishes between individual opinions and collective sentiments when aggregating views from multiple users. Confirm that referenced features are associated with their correct corresponding entities.
\\\\
\textbf{(4) Check Sentiment Consistency}
\\
Check for sentiment polarity errors, particularly in cases involving irony, sarcasm, or nuanced expressions. Ensure that positive opinions are not mistakenly presented as negative and vice versa. Verify that the insight accurately captures the tone and emotional valence of the original opinions, including subtle sentiment expressions that may be context-dependent. Confirm that intensity modifiers (e.g., "very," "somewhat," "extremely") are appropriately preserved when they significantly impact the expressed opinion.
\\\\
\textbf{(5) Verify Three-Line Structure}

\textbf{Line 1 (Frequent or Emphasized Features) Verification}
\begin{itemize}[leftmargin=*,topsep=4pt,itemsep=4pt,parsep=0pt]
    \item Confirm that the insight accurately identifies and highlights features most frequently mentioned or emphasized in the original text 
    \item Check frequency counts to verify that truly common themes are included in the insight
    \item Check if minority opinions aren't overrepresented or majority opinions underrepresented  
\end{itemize}

\textbf{Line 2 (Strongly Praised or Criticized Features) Verification}
\begin{itemize}[leftmargin=*,topsep=4pt,itemsep=4pt,parsep=0pt]
    \item Confirm the insight clearly identifies features that received particularly strong praise or criticism 
    \item Verify that the intensity of opinions is accurately conveyed (using appropriate intensity indicators) 
    \item Check that the distinction between mild opinions and strong sentiments is preserved 
\end{itemize}

\textbf{Line 3 (Conclusion/Key Takeaway) Verification}
\begin{itemize}[leftmargin=*,topsep=4pt,itemsep=4pt,parsep=0pt]
    \item Confirm the presence of a concise conclusion that synthesizes the overall sentiment 
    \item Verify this conclusion accurately reflects the predominant message across all original comments 
    \item Check that the conclusion doesn't introduce new information not supported by the original text 
\end{itemize}

\end{minipage} \\ 
\bottomrule
\end{tabular}
\caption{\proposed: Opinion-centric insight annotation and verification process.}
\label{tab:summary_guideline}
\end{table*}

%% file: TAB/086summary_case_study.tex
\begin{table*}[t]
  \centering
  \small
  \begin{tabular}{@{} p{0.60\textwidth}  p{0.30\textwidth} @{}}
    \toprule
    \textbf{Input Content} & \textbf{Opinion-centric Insight} \\
    \midrule
    \begin{minipage}[t]{\linewidth}
      \textcolor{teal}{\textbf{[YouTube]}} \\
      \textbf{Title:} Here’s Why Everyone Hates the Mercedes-AMG GLE63 Coupe\\[0.5ex]
      \textbf{Post \#1 (Person 1):} I actually like the ways these look. They look big and aggressive\\[0.5ex]
      \textbf{Comment \#1 (Person 2):} Said no one ever\\[0.5ex]
      \textbf{Comment \#2 (Person 3):} not everyone has good taste. I dont like it because it looks good but the coupe roof line, and that’s the point of the car. So it’s like it could be perfect but it’s not, to me at least. Looks wise. They should’ve changed the front too.\\[0.5ex]
      \textbf{Comment \#3 (Person 4):} no they look obnoxious and awkwardly proportioned. you’d have to be insane to spend over \$10K more for a less efficient, less practical version of a much better car in every aspect.\\[0.5ex]
      \textbf{Comment \#4 (Person 5):} If u like dome shaped overpriced…\\[0.5ex]
      \textbf{Comment \#5 (Person 6):} It’s a fat sedan 
    \end{minipage}
    &
    \begin{minipage}[t]{\linewidth}
      The Mercedes-AMG GLE63 Coupe receives mixed reviews, with users criticizing its design as awkwardly proportioned, less efficient, and less practical compared to other models. Some find the coupe's roofline and overall look obnoxious and overpriced, while a minority appreciate its big and aggressive appearance. Overall, the sentiment leans negative, with the car's aesthetics and practicality being the main points of contention.
    \end{minipage}
    \\
    \midrule
    \begin{minipage}[t]{\linewidth}
    \textcolor{teal}{\textbf{[Reddit]}}\\[0.5ex]
    \textbf{Title}: What are some versions/generations of cars that have been mostly under the radar?\\[0.5ex]
    \textbf{Post \#11518 (Author: Person 1):} Y34 Infiniti M45.340hp V8 luxury sedan with strong, square styling. Roughly 9,000 imported over the two years on sale.Parts availability is a bitch.\\[0.5ex]
    \textbf{Comment \#11519 (Author: Person 2, Reply to reply comment \#11518):} Good choice. An actual car that people dont talk about. There was also the M56 with a VK56 V8 that made 420 horsepower. Also rare. Heard one with an exhaust a while ago and it sounded pretty good.\\[0.5ex]
    \textbf{Comment \#11520 (Author: Person 3, Reply to reply comment \#11519):} Yeah, the M56 is really cool and almost never gets talked about. Infiniti also continued the V8 when they renamed the M37M56 to the Q70, but a V8 Q70 is even rarer than an M56.\\[0.5ex]
    \textbf{Comment \#11521 (Author: Person 4, Reply to reply comment \#11518):} And they look so rad. Quirky Japanese styling that wasnt super well received by the western market at the time, but few things look as good as a murdered out M45\\[0.5ex]
    ...
    \end{minipage}
    &
    \begin{minipage}[t]{\linewidth}
    The Y34 Infiniti M45 is celebrated for its strong, square styling and potential for timeless appeal with the right modifications, though parts availability is a challenge. Its quirky Japanese design was not initially well-received in the Western market, yet it is now appreciated for its unique and sinister look, comparable to the w109 Mercedes. The M56 variant, with its 420 horsepower V8, is noted for its rarity and impressive sound, while the V8 Q70 is even rarer, adding to the allure of these distinctive models.
    \end{minipage}
    \\
    \midrule
    \begin{minipage}[t]{\linewidth}
    \textcolor{teal}{\textbf{[Reveiw Site]}}\\[0.5ex]
    \textbf{Title}: Turn OFF Automatic Software Updates\\[0.5ex]
    Took awhile but have grown from tolerating to appreciating this SUV. 
    Replacement of a defective 12v battery and the dealer updating my car from the 2.0 to the 2.1 software resolved the initial issues that initially were bricking my XC60 B5.  
    Second, I turned OFF the automatic software updates which you can do in check for software update submenu. Sticking with version 2.1 for now since it has made my car stable (at least on my B5 trim).  
    2.2 is available which includes Apple CarPlay but that version was JUST released and the initial bugs have not been worked out. 
    So buggy that 2.2 is not available for an OTA update specifically for B5 trims since it causes known problems. 
    ... 
    \end{minipage}
    &
    \begin{minipage}[t]{\linewidth}
    The Volvo XC60 B5 receives mixed feedback, with users appreciating its comfort and unique design, particularly in the R-design trim. However, software issues are a significant concern, as the latest 2.2 version, which includes Apple CarPlay, is buggy and unavailable for OTA updates on B5 trims. Users recommend sticking with the stable 2.1 version and turning off automatic updates to maintain vehicle stability, while the native Google OS is praised for its functionality.
    \end{minipage}
    \\
    \midrule
    \begin{minipage}[t]{\linewidth}
    \textcolor{teal}{\textbf{[Blog]}}\\[0.5ex]
    \textbf{Title}: The Little Things: 2024 Hyundai Kona’s Heat Buttons Are Oddly Satisfying\\[0.5ex]
    The Hyundai Kona has a very open and room-like interior, with some well-placed buttons.
    I’ve been on an ergonomics kick lately, closely scrutinizing the button placement and cockpit layouts of every car I get into. The 2024 Hyundai Kona has a uniquely open and homey-feeling interior, but one aspect of it that struck me was how perfect the heated seat and steering wheel buttons are.
    Before your eyes roll out of your head at the thought of praising something as mundane as a button, consider this: Heated seat and steering wheel controls are features you might use every day, multiple times a day, for months and months while you own a car. 
    ...
    \end{minipage}
    &
    \begin{minipage}[t]{\linewidth}
    The 2024 Hyundai Kona is praised for its interior design, featuring a uniquely open and homey feel with well-placed, physical buttons that enhance user engagement and ease of use. The heated seat and steering wheel controls receive high marks for their perfect placement and functionality, including a rare two-level heat setting for the steering wheel. Despite some criticism of its front-end design and limited acceleration, the overall sentiment is positive, highlighting the Kona's thoughtful ergonomic touches and practical features.
    \end{minipage}
    \\
    \bottomrule
  \end{tabular}
  \caption{Examples of \proposed opinion-centric-insights by content type}
  \label{tab:oig_gold_summary_case_study2}
\end{table*}

%% file: TAB/086summary_case_study_g_eval.tex
\begin{table*}
  \centering
  \small
  \begin{tabularx}{\textwidth}{%
    >{\arraybackslash}p{0.15\textwidth}
    >{\arraybackslash}X
  }
    \toprule
    \multicolumn{2}{l}{\textbf{Case Study: Review Site (1/2)}} \\
    \midrule
    \textbf{Input Content}
      & 
      \textbf{Title}: Turn OFF Automatic Software Updates
      
      Took awhile but have grown from tolerating to appreciating this SUV.  
      Replacement of a defective 12v battery and the dealer updating my car from the 2.0 to the 2.1 software resolved the initial issues that initially were bricking my XC60 B5.  
      Second, I turned OFF the automatic software updates which you can do in check for software update submenu. 
      Sticking with version 2.1 for now since it has made my car stable (at least on my B5 trim).  2.2 is available which includes Apple CarPlay but that version was JUST released and the initial bugs have not been worked out. 
      So buggy that 2.2 is not available for an OTA update specifically for B5 trims since it causes known problems. While CarPlay is in high demand the native Google OS has been great and the Google map appears on the driver display which CarPlay currently wont. 
      That said, recommend finding the software  version that enables stability for your car/trim and then turn off automatic updates in the settings. 
      For my XC60 B5 I am going to stick with 2.1 until forced to update or until the blogs indicate Volvo and Google are aligning better. 
      Aside from the software gremlins this SUV is comfortable, super handsome in my r-design trim, and feels a touch unique.
    \\ 
    \midrule
    \textbf{Model Generated Insight}
      & The reviewer initially faced issues with their Volvo XC60 B5 due to a defective 12v battery and problematic software updates, but these were resolved by upgrading to software version 2.1. 
      They advise against automatic software updates, as version 2.2, though offering Apple CarPlay, is buggy and not recommended for B5 trims. 
      The reviewer appreciates the native Google OS, especially its integration with the driver display, and plans to stick with version 2.1 for stability. 
      Despite software challenges, they find the SUV comfortable, stylish in the R-design trim, and unique.  
    \\
    \midrule
    \textbf{Gold Insight}
      & 
      The Volvo XC60 B5 receives mixed feedback, with users appreciating its comfort and unique design, particularly in the R-design trim. 
      However, software issues are a significant concern, as the latest 2.2 version, which includes Apple CarPlay, is buggy and unavailable for OTA updates on B5 trims. 
      Users recommend sticking with the stable 2.1 version and turning off automatic updates to maintain vehicle stability, while the native Google OS is praised for its functionality. 
    \\
    \midrule
    \textbf{A3CU}
      & 60.91
    \\
    \bottomrule
  \end{tabularx}
  \caption{Case Study (1) of LLM-Judge Evaluation for the OIG Task. Opinion-centric insight generated by \gptfouro.}
  \label{tab:llm_judge_eval_case_study_11}
\end{table*}

\begin{table*}
  \centering
  \small
  \begin{tabularx}{\textwidth}{%
    >{\arraybackslash}p{0.15\textwidth}
    >{\arraybackslash}X
  }
    \toprule
    \multicolumn{2}{l}{\textbf{Case Study: Reddit (1/2)}} \\
    \midrule
    \textbf{Input Content}
      & \textbf{Title}: Ford starts deliveries of enhancedrange etransit
    
    \textbf{Comment 35276 (Author: Person 1)}: hate the trend of just using an ICE hull for an EV, so uninteresting and ugly. EV allows you to change the form factor of cars. I like new startups that make interesting and practical looking cars.

    \textbf{Comment 35277 (Author: Person 2, Reply to reply comment \#35276)}: Its a work van. Fleets like having an already existing supply of commonly used and broken parts much more than they like having pretty looking vans.

    \textbf{Comment 35278 (Author: Person 3, Reply to reply comment \#35276)}: Yeah but cars are the way they are for a lot of other reasons. And it actually makes a ton of financial sense. Some of the hypothetical benefits are not really as beneficial when insurance, repair, maintenance, learning curve for mechanics are taken into consideration.

    \textbf{Comment 35279 (Author: Person 4, Reply to reply comment \#35276)}: Vans are gonna be fuckin boxes no matter what, its the point
    \\ 
    \midrule
    \textbf{Model Generated Insight}
      & The review discussion centers around the design and practicality of electric vehicles (EVs) using traditional internal combustion engine (ICE) designs. 
      Person 1 expresses dissatisfaction with the trend of using ICE hulls for EVs, finding them uninteresting and aesthetically displeasing, and prefers innovative designs from new startups. 
      In contrast, Person 2 argues that for work vans, practicality and the availability of parts are more important than aesthetics. 
      Person 3 supports this view, highlighting the financial and practical benefits of using existing designs, considering factors like insurance, repair, and maintenance. 
      Person 4 bluntly states that vans are inherently boxy due to their functional purpose, implying that design changes are limited by their utility.
    \\
    \midrule
    \textbf{Gold Insight}
      & The Ford e-Transit receives mixed feedback, with some users criticizing its design as uninteresting and ugly due to its use of an ICE hull, while others appreciate the practicality and financial sense of maintaining a familiar form. The vehicle's reliance on commonly used parts is seen as beneficial for fleet maintenance, despite the lack of aesthetic appeal. Overall, the sentiment reflects a divide between those valuing innovative design and those prioritizing practicality and cost-effectiveness.
    \\
    \midrule
    \textbf{A3CU}
      & 10.12
    \\
    \bottomrule
  \end{tabularx}
  \caption{Case Study (2) of LLM-Judge Evaluation for the OIG Task. Opinion-centric insight generated by \gptfouro.}
  \label{tab:llm_judge_eval_case_study_21}
\end{table*}

\begin{table*}
  \centering
  \small
  \begin{tabularx}{\textwidth}{%
    >{\arraybackslash}p{0.15\textwidth}
    >{\arraybackslash}X
  }
    \toprule
    \multicolumn{2}{l}{\textbf{Case Study: Review Site (2/2)}} \\
    \midrule
    \textcolor{teal}{\textbf{Faithfulness}}: 5
      & 
    \textbf{Accuracy of Software and Battery Issues}: 
    The insight summary correctly notes the initial issues the user faced due to a defective 12v battery and the problematic software updates, which closely aligns with what the review text describes. 
    It also correctly mentions how these issues were largely resolved by upgrading to version 2.1.

    \textbf{Software Updates and Recommendations}: 
    The insight summary accurately captures the user's advice against automatic software updates, specifically highlighting that version 2.2 is buggy, particularly for the B5 trim, which is consistent with the original review's observations. 
    It also correctly represents the reviewer's intent to stick with version 2.1 for stability. \\ 
    \midrule
    \textcolor{teal}{\textbf{Coverage}}: 4
      & \textbf{Comprehensive Coverage of Main Issues}: 
      The insight summary effectively highlights the initial issues faced by the reviewer, namely the defective 12v battery and problematic software updates, and how these were resolved by upgrading to software version 2.1. ...

    \textbf{Inclusion of Recommendation and Software Concerns}: 
    The insight summary correctly encapsulates the reviewer's advice against automatic updates due to the issues with version 2.2, emphasizing its buggy nature and incompatibility with B5 trims. ...

    \textbf{Minor Missing Details}: 
    Although the insight summary is generally accurate, it doesn't mention the specific aspect of finding a stable software version and then turning off automatic updates, which adds a nuanced detail to the reviewer's overall advice. ...
    \\
    \midrule
    \textcolor{teal}{\textbf{Specificity}}: 4
      & \textbf{Resolution Details}: 
      The insight summary covers the resolution process involving the upgrade to software version 2.1 to address initial vehicle issues. ...

    \textbf{Specific Features and Versions}: 
    The insight summary specifies the difference between software versions 2.1 and 2.2, noting the inclusion of Apple CarPlay in version 2.2. ...

    \textbf{General Impression and Recommendations}: 
    While largely specific in detail, certain generalizations are made, such as mentioning the comfort and uniqueness of the SUV without referencing specific features that contribute to these impressions. ...
    \\
    \midrule
    \textcolor{teal}{\textbf{Insightfulness}}: 4
      & \textbf{Comprehensive Coverage}: 
      The insight summary effectively covers the key points of the review, including the issue with the battery, the software problems, and the resolution through the 2.1 software update. ...

    \textbf{Value-Added Advice}: 
    By recommending a strategic approach to software updates and emphasizing the stability offered by version 2.1, the insight summary provides actionable insights that can directly inform decision-making for potential users of the XC60 B5, particularly those concerned about software reliability.

    \textbf{Balanced Perspective}: 
    The insight summary maintains a balance by acknowledging the non-software-related positives of the vehicle, such as comfort and aesthetic appeal, which enriches understanding for readers interested in a broader evaluation of the SUV's attributes. ...
    \\
    \midrule
    \textcolor{teal}{\textbf{Intent}}: 5
      & \textbf{Tone and Intent Preservation}: 
      The insight summary effectively captures the reviewer's tone of cautious appreciation towards the vehicle, highlighting the resolve to prioritize stability over newer, but potentially problematic, software updates. ...

    \textbf{Emotional Nuances}: 
    The emotional essence of the original review, which oscillates between initial frustration and eventual appreciation, is well-preserved. ...

    \textbf{Detail and Specificity}: 
    Key details concerning the software update issues, particularly the problems associated with version 2.2 and the decision to stick with version 2.1, are thoroughly covered in the insight summary. ...
    \\
    \midrule
    \textcolor{teal}{\textbf{Fluency}}: 5
      & \textbf{Natural Flow and Language Usage}: 
      The insight summary is written in a manner that mimics a natural and conversational tone. 
      It concisely recounts the original review's main points without introducing excessive technical jargon or unnecessary detail.

    \textbf{Grammar and Sentence Structure}: 
    The insight summary is void of grammatical errors or awkward phrasing, making it easy for the reader to follow. 
    Sentences are constructed well, contributing to a smooth reading experience.

    \textbf{Clarity and Cohesion}: 
    Each point in the insight summary builds logically upon the previous, maintaining clear cohesion throughout. 
    This makes it accessible even to readers unfamiliar with the context of the review text. \\
    \bottomrule
  \end{tabularx}
  \caption{Case Study of LLM-Judge Evaluation for the OIG Task. Opinion-centric insight generated by \gptfouro.}
  \label{tab:llm_judge_eval_case_study_12}
\end{table*}

\begin{table*}
  \centering
  \small
  \begin{tabularx}{\textwidth}{%
    >{\arraybackslash}p{0.15\textwidth}
    >{\arraybackslash}X
  }
    \toprule
    \multicolumn{2}{l}{\textbf{Case Study: Reddit (2/2)}} \\
    \midrule
    \textcolor{teal}{\textbf{Faithfulness}}: 5
      & 
    \textbf{Accurate Representation}: 
    The insight summary accurately captures the main points and perspectives shared by the commenters without distorting their meaning. ...
    
    \textbf{No Omitted Key Details}: 
    The core opinions about the use of ICE hulls for EVs, the practical considerations for work vans, and the limited design flexibility due to utility needs are all present in the insight summary. 
    This shows no significant omissions.
    
    \textbf{Absence of Hallucinated Information}: 
    The insight summary does not introduce any information that was not present in the original text, ensuring that it stays true to the opinions and insights provided by the commenters. ...
    \\ 
    \midrule
    \textcolor{teal}{\textbf{Coverage}}: 5
    &
    \textbf{Comprehensive Representation}: The insight summary thoroughly captures the primary opinions expressed by each participant in the review text. 
    It details Person 1's dislike for using ICE hulls in EVs, Person 2's focus on practicality and cost-effectiveness discussed by Person 2, Person 3's considerations of financial benefits, and Person 4's functional view of van design. 
    
    \textbf{Inclusion of Key Points}: No significant opinions are omitted. The contrasting views regarding aesthetic values versus practical functionality and financial considerations are clearly highlighted, showcasing the diverse perspectives discussed in the review.
    
    \textbf{Conciseness and Clarity}: The insight summary provides a concise yet clear representation of the review's key discussions, effectively relaying the main points without unnecessary details, ensuring all significant opinions are well-covered. \\
    \midrule
    \textcolor{teal}{\textbf{Specificity}}: 5
    &
    \textbf{Concrete Alignment with Comments}: The insight summary effectively captures the conversation between different users by directly referencing the points they make. 
    It notes Person 1's dissatisfaction with ICE-based EV designs and preference for innovative startups, Person 2's emphasis on practicality and part availability for work vans, Person 3's mention of financial and practical factors, and Person 4's blunt assertion about vans' inherent boxiness. ...
    
    \textbf{No Overgeneralization}: 
    At no point does the insight summary veer into vague or generic statements. 
    Each opinion is not only linked to the respective commenter but also articulated with clear explanation of their rationale, demonstrating a detailed and precise understanding of the user opinions.
    
    \textbf{Point-wise Clarity}: 
    The insight summary maintains a structurally coherent format by presenting each perspective distinctly. ...
    \\
    \midrule
    \textcolor{teal}{\textbf{Insightfulness}}: 4
    &
    \textbf{Comprehensive Overview}: The insight summary effectively synthesizes various perspectives from the commentary, covering not only the dissatisfaction with traditional designs from Person 1 but also contrasting practical views from others who value utility over aesthetics. ...
    
    \textbf{Emphasis on Practical Factors}: By highlighting the points about practicality—such as access to parts, financial logic, and the vans' purpose—the insight summary goes beyond mere restatement of comments and offers a deeper understanding of the practical considerations influencing design decisions in the EV market.
    
    \textbf{Concludes with Functional Insight}: The insight summary concludes with Person 4's point about the inherent boxy design of vans due to their function, driving home the idea that practicality often trumps aesthetic innovation in vehicle design. ...
    \\
    \midrule
    \textcolor{teal}{\textbf{Intent}}: 5
    &
    \textbf{Tone and Intent Preservation}: The insight summary accurately captures the tone and intent of each comment within the discussion. Person 1's dissatisfaction with the trend is clearly articulated, as is the practical-focused tone of Persons 2, 3, and 4. ...
    
   \textbf{ Nuance and Specificity}: The insight summary encapsulates the nuances in each participant's perspective. 
   It highlights Person 1's preference for innovative design, contrasts this with the practical considerations highlighted by Person 2 and Person 3, and points out Person 4's straightforward viewpoint regarding the inherent functionality of vans.
   
   \textbf{Consistent Style and Balance}: The insight summary maintains a balanced view and communicates both sides of the argument without straying from the original intent of the discourse. ...
    \\
    \midrule
    \textcolor{teal}{\textbf{Fluency}}: 5
    &
    \textbf{Grammar and Syntax}: The insight summary is grammatically sound, with proper sentence structures, clear subject-verb agreement, and correct use of punctuation, enhancing readability.
    
    \textbf{Flow and Coherence}: The transition between points expressed by different persons is seamless, logically structured, allowing for a coherent understanding of conflicting viewpoints on EV and ICE designs.
    
    \textbf{Clarity and Readability}: The language used is straightforward and easy to follow, with precise vocabulary choices that appropriately convey complex opinions in an accessible manner.
    \\
    \bottomrule
  \end{tabularx}
  \caption{Case Study of LLM-Judge Evaluation for the OIG Task. Opinion-centric insight generated by \gptfouro.}
  \label{tab:llm_judge_eval_case_study_22}
\end{table*}

%% file: TAB/089annotation_ui.tex
\begin{figure*}
    \centering
    \includegraphics[width=0.80\textwidth]{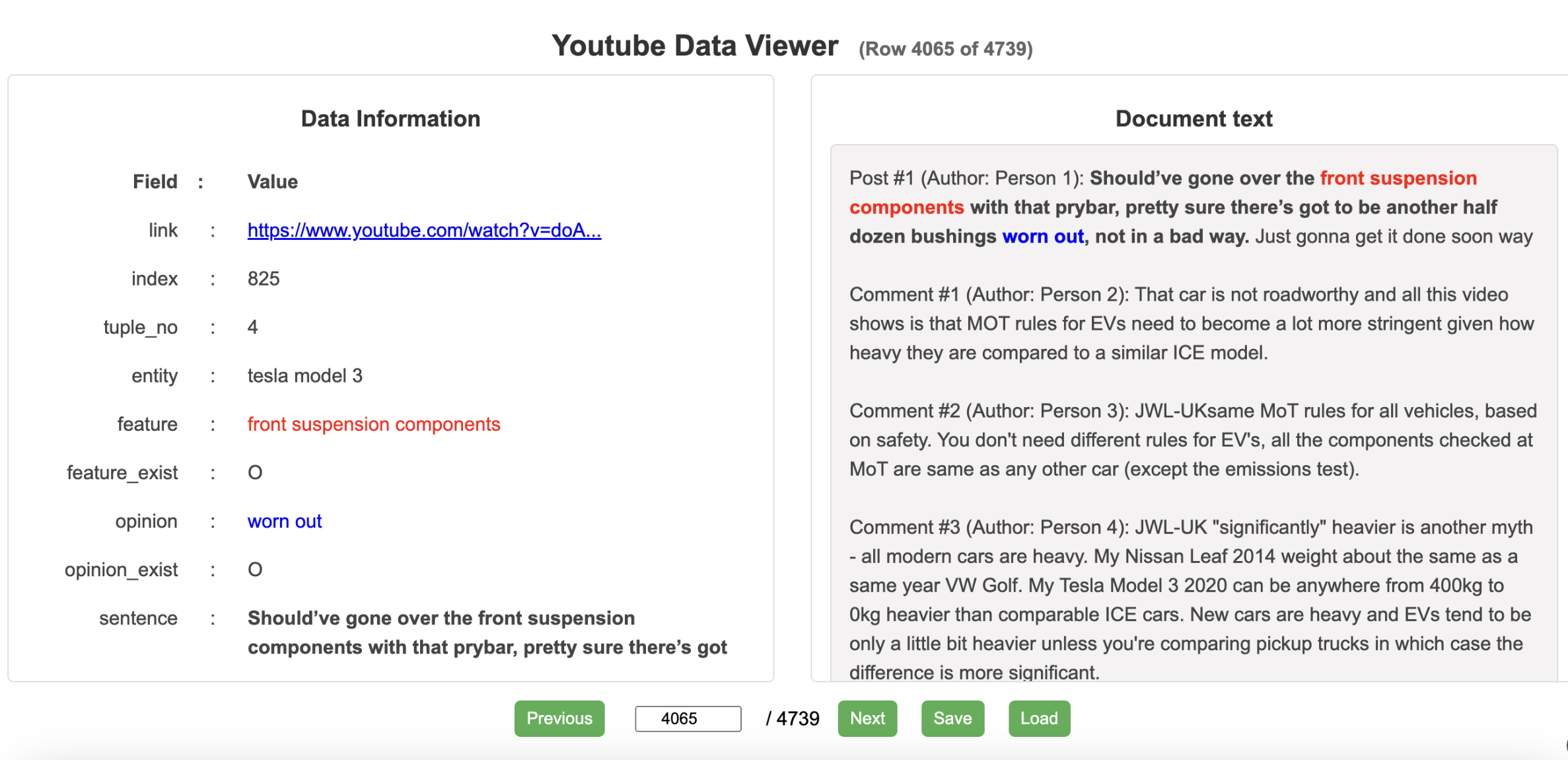}
    \caption{\proposed – Annotation UI used for Entity-feature-opinion tuple set annotation. This example shows a case where both the feature and opinion are present in the content.}
    \label{fig:tuple_ui_1}
\end{figure*}
\begin{figure*}
    \centering
    \includegraphics[width=0.80\textwidth]{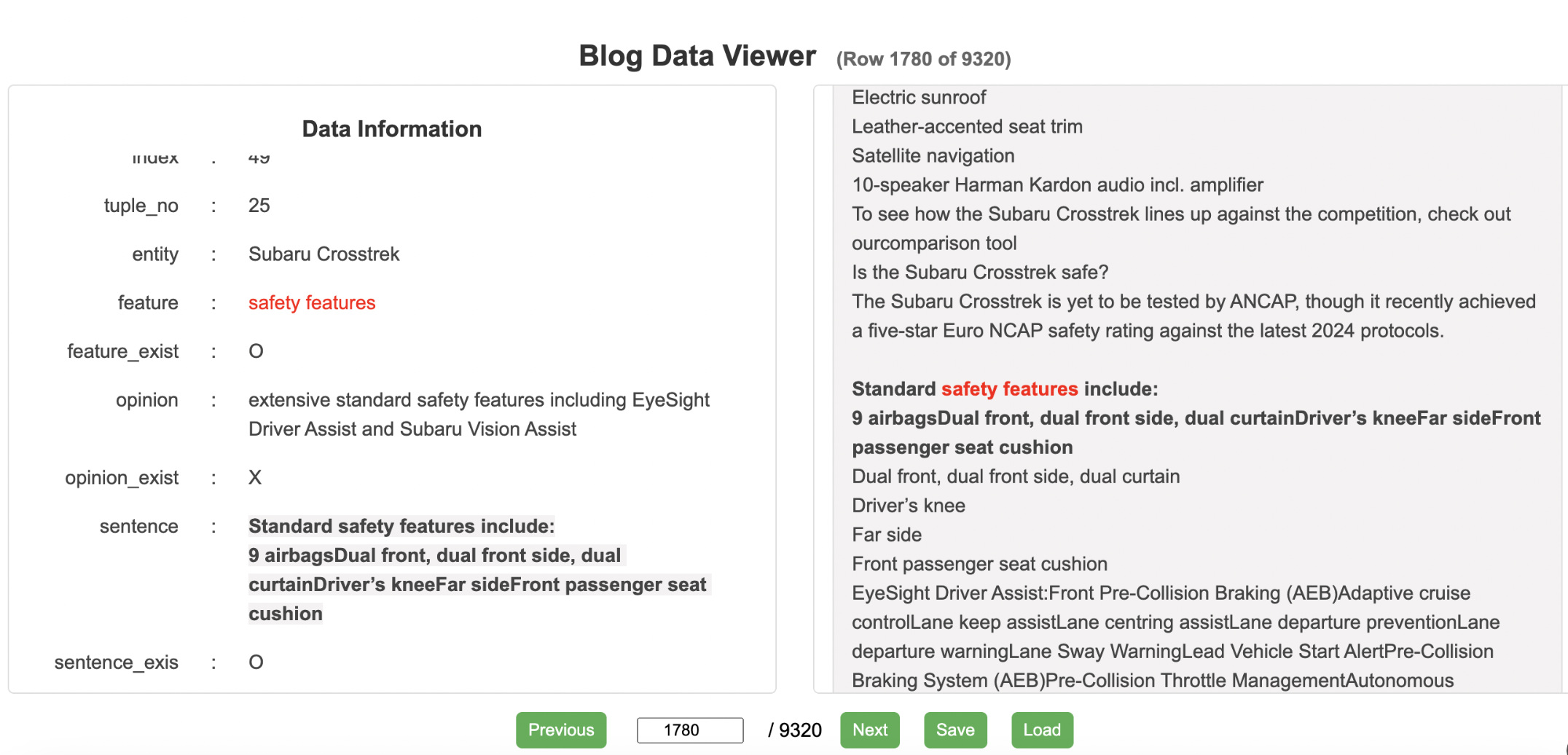}
    \caption{\proposed – Annotation UI used for Entity-feature-opinion tuple set annotation. This example shows a case where the feature is present in the content, but the opinion is not.}
    \label{fig:tuple_ui_2}
\end{figure*}

\begin{figure*}
    \centering
    \includegraphics[width=0.80\textwidth]{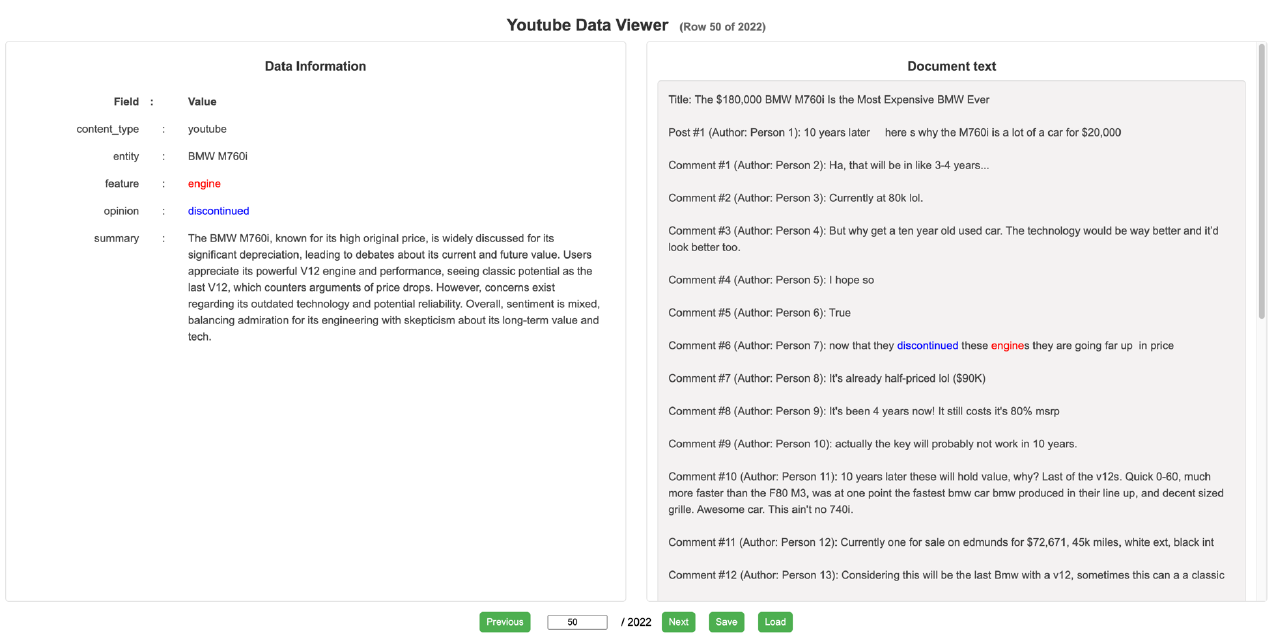}
    \caption{\proposed – Annotation UI used for opinion-centric insight annotation.}
    \label{fig:summary_ui}
\end{figure*}

%% file: TAB/088human_eval_amt.tex
\begin{figure*}[h]
\centering
\includegraphics[width=0.90\textwidth]{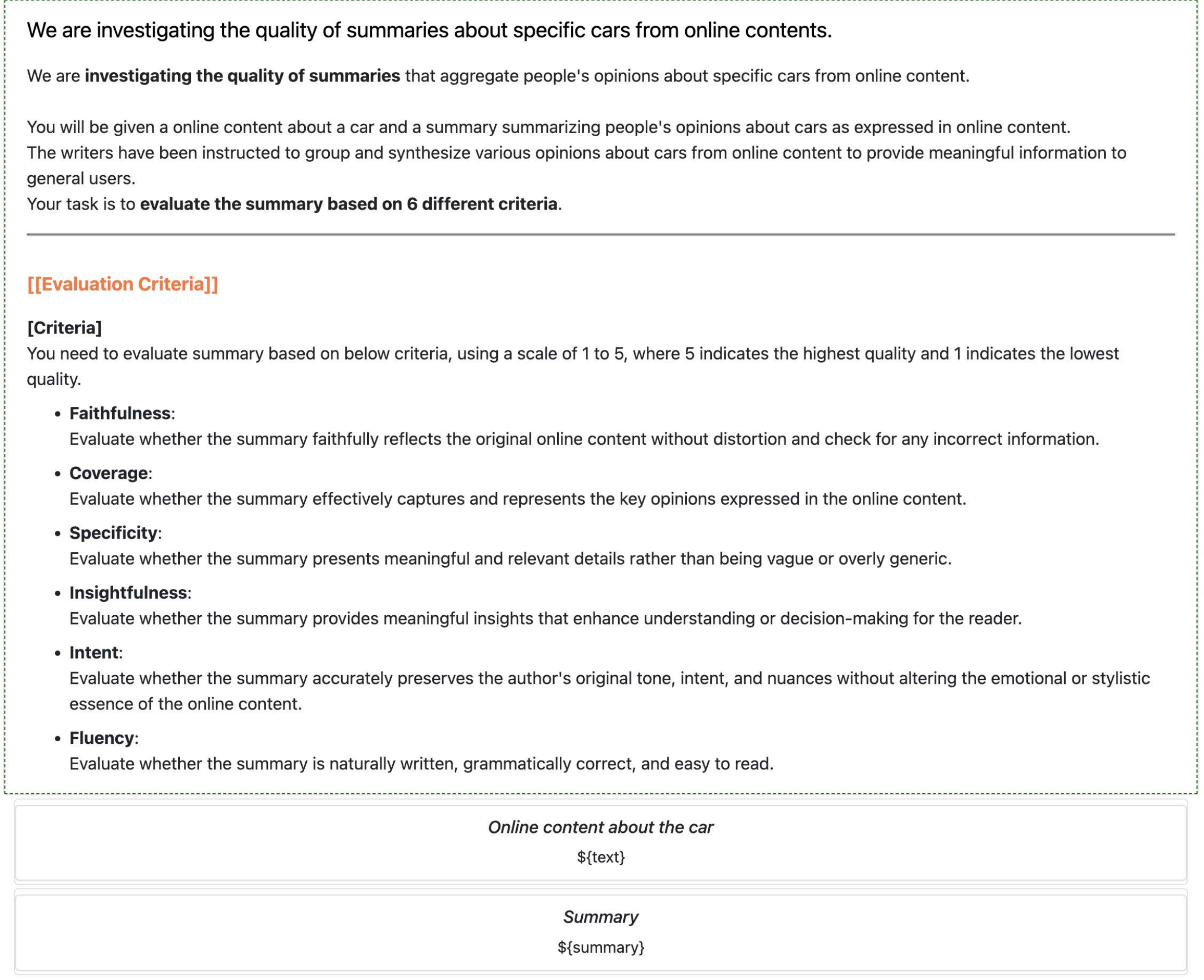}
\caption{The interface for human evaluation (Instruction part).}
\label{fig:humaneval_main}
\end{figure*}

\begin{figure*}[h]
\centering
\includegraphics[width=\textwidth]{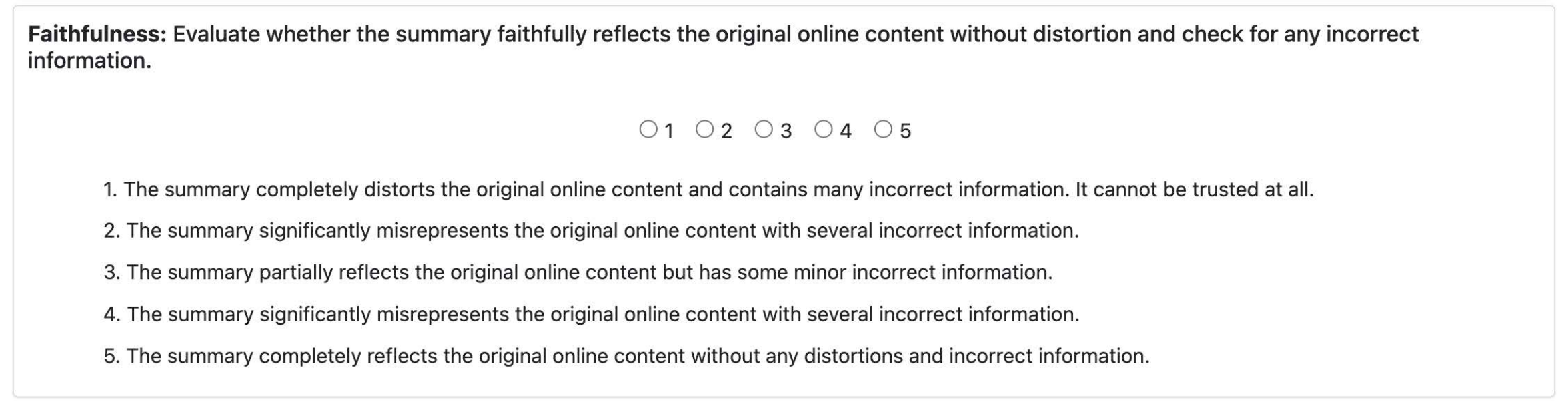}
\caption{The interface for human evaluation (Faithfulness).}
\label{fig:humaneval_faithfulness}
\end{figure*}

\begin{figure*}[h]
\centering
\includegraphics[width=\textwidth]{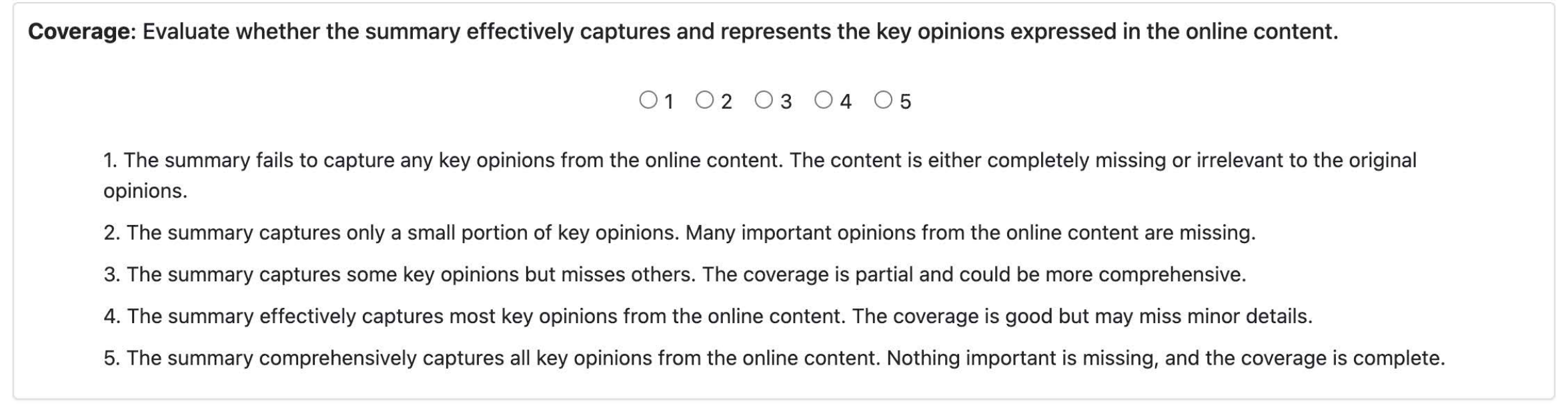}
\caption{The interface for human evaluation (Coverage).}
\label{fig:humaneval_coverage}
\end{figure*}

\begin{figure*}[h]
\centering
\includegraphics[width=\textwidth]{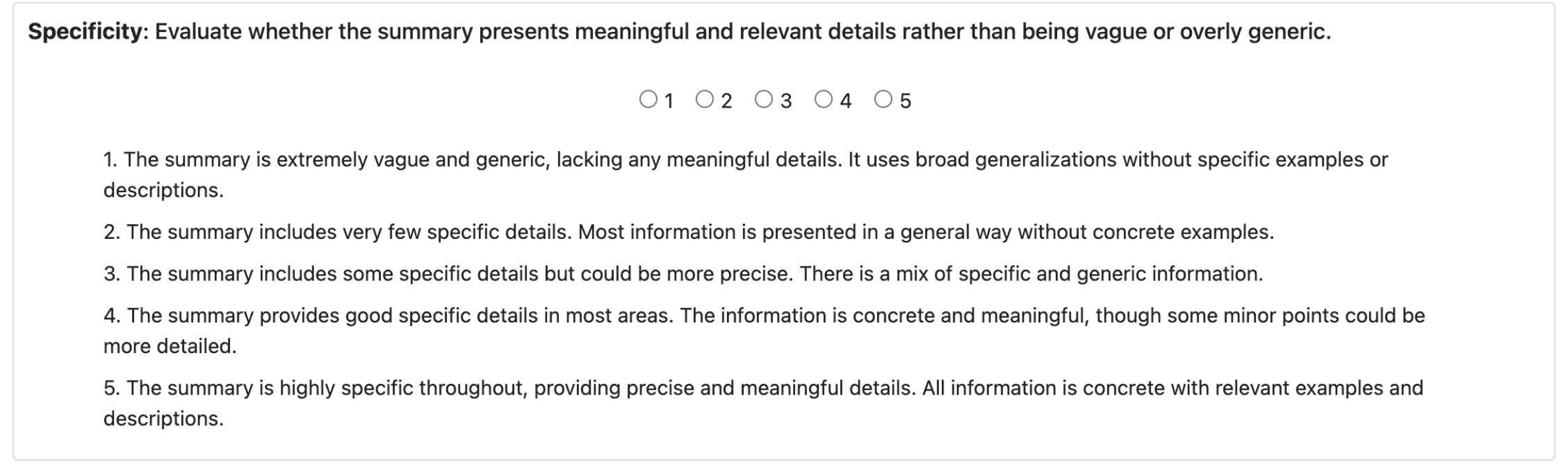}
\caption{The interface for human evaluation (Specificity).}
\label{fig:humaneval_specificity}
\end{figure*}

\begin{figure*}[h]
\centering
\includegraphics[width=\textwidth]{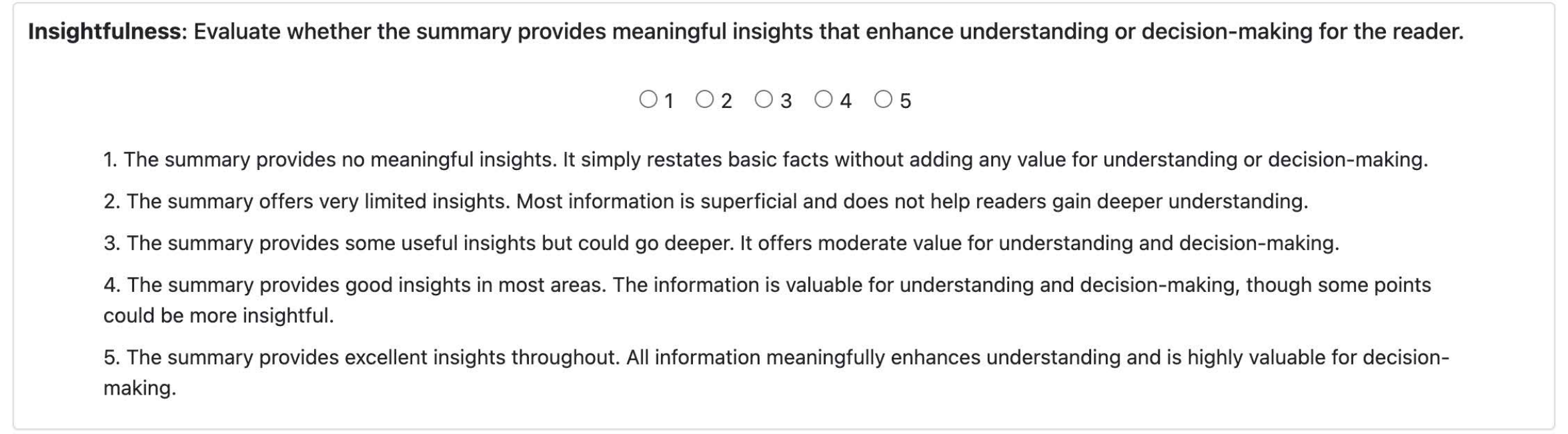}
\caption{The interface for human evaluation (Insightfulness).}
\label{fig:humaneval_insightfulness}
\end{figure*}

\begin{figure*}[h]
\centering
\includegraphics[width=\textwidth]{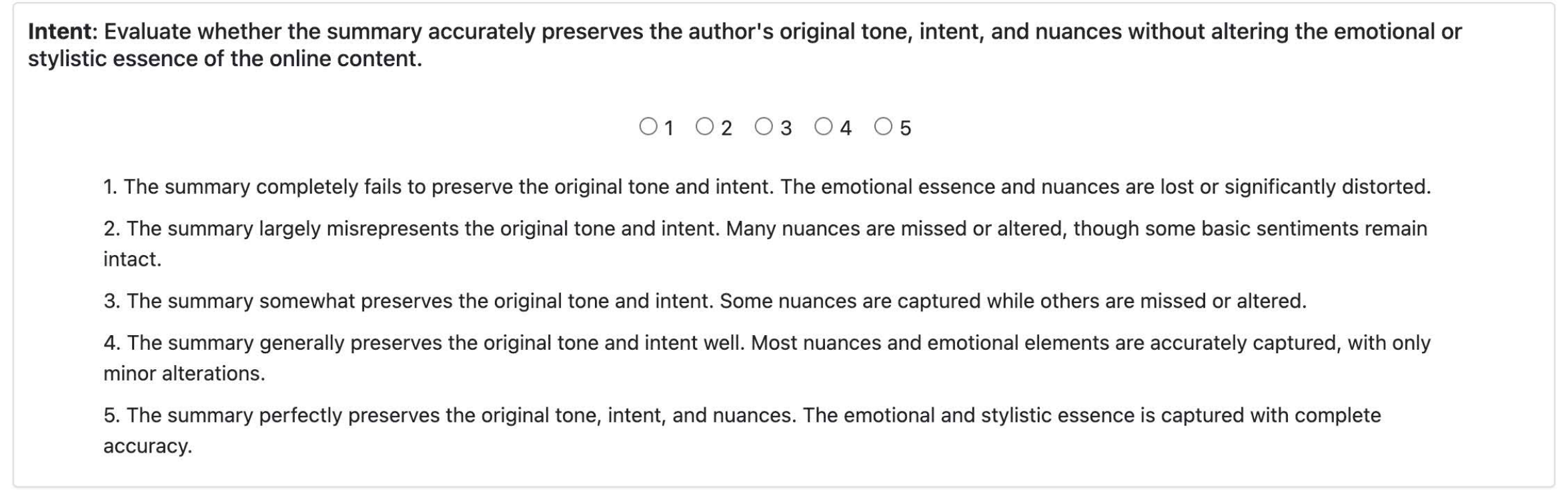}
\caption{The interface for human evaluation (Intent).}
\label{fig:humaneval_intent}
\end{figure*}

\begin{figure*}[h]
\centering
\includegraphics[width=\textwidth]{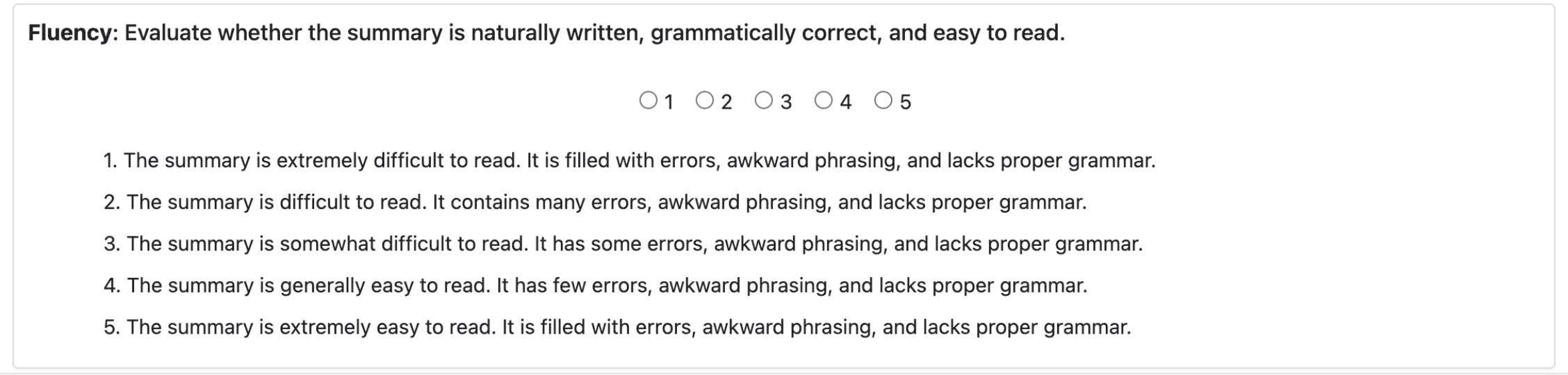}
\caption{The interface for human evaluation (Fluency).}
\label{fig:humaneval_fluency}
\end{figure*}